\documentclass{article} %

\usepackage{iclr2026_conference,times}

\newif\ifpreprint
\preprinttrue  %

\ifpreprint\iclrfinalcopy\fi

\usepackage{hyperref}
\usepackage{url}
\usepackage{graphicx}
\usepackage{booktabs}
\usepackage{multirow}
\usepackage{colortbl}
\usepackage{wrapfig}
\usepackage{subcaption}
\usepackage{float}
\usepackage{pgfplots}
\usepackage{enumitem}
\usetikzlibrary{arrows.meta,decorations.pathreplacing}
\pgfplotsset{compat=1.18}
\usepgfplotslibrary{groupplots}
\usepackage{pgfplotstable}
\usepackage{xcolor}
\usepackage{amssymb, amsmath}

\definecolor{bestrow}{HTML}{D5E8D4}
\definecolor{mcA}{HTML}{2B4A7C}   %
\definecolor{mcB}{HTML}{5B8DB8}   %
\definecolor{mcC}{HTML}{C0392B}   %
\definecolor{mcD}{HTML}{E67E73}   %
\definecolor{mcE}{HTML}{27864E}   %
\definecolor{mcF}{HTML}{7DC29A}   %
\definecolor{urlred}{HTML}{D00000} %
\hypersetup{urlbordercolor={1 1 1}}

\usepackage{cleveref}
\usepackage{etoc}
\crefname{section}{sec.}{secs.}
\Crefname{section}{Sec.}{Secs.} %
\crefname{equation}{eq.}{eq.}
\Crefname{eqation}{Eq.}{Eqs.} %

\newcommand{\ours}{\textsc{Flex-$\pi$}}

\definecolor{bestrow}{rgb}{.835,.91,.831}

\definecolor{darkyellow}{rgb}{0.7,0.5,0.0}
\definecolor{gray}{rgb}{0.62352941176,0.63137254902,0.6431372549}

\newcommand{\dino}{d}
\newcommand{\point}{p}
\newcommand{\enc}{\mathrm{Enc}}
\newcommand{\dec}{\mathrm{Dec}}
\newcommand{\dinoenc}{\mathrm{DINO}}

\definecolor{ourmethod}{HTML}{2e7d3e} %
\definecolor{pi}{HTML}{717171} 
\definecolor{maniflow}{HTML}{7fb3d5} 
\definecolor{fastwam}{HTML}{e0a526}    %
\definecolor{lingbot}{HTML}{7a93a8}    %

\newcommand{\ourcolor}[1]{\textcolor{ourmethod}{#1}}

\newcommand{\ourexp}[1]{\textbf{\textcolor{ourmethod}{\ours{}}}}
\newcommand{\fastwam}{\textbf{\textcolor{fastwam}{Fast-WAM}}}
\newcommand{\pibaseline}{\textcolor{pi}{$\mathbf{\pi_{0.5}}$}}
\newcommand{\maniflow}{\textbf{\textcolor{maniflow}{ManiFlow}}}
\newcommand{\lingbot}{\textbf{\textcolor{lingbot}{LingBot-VA}}}

\newcommand{\algcomment}[1]{\textcolor{gray}{#1}}
\newcommand{\latentmodel}{v_\theta}
\newcommand{\inputmask}{\mathbf{m}^\text{in}}
\newcommand{\outputmask}{\mathbf{m}^\text{out}}

\definecolor{masks}{HTML}{E59EDD}
\definecolor{mot}{HTML}{9FCBDA}
\definecolor{conditioning}{HTML}{E8E3D8}
\definecolor{mcA}{HTML}{2B4A7C}   %
\definecolor{mcB}{HTML}{5B8DB8}   %
\definecolor{mcC}{HTML}{C0392B}   %
\definecolor{mcD}{HTML}{E67E73}   %
\definecolor{mcE}{HTML}{27864E}   %
\definecolor{mcF}{HTML}{7DC29A}   %
\definecolor{bestrow}{HTML}{D5E8D4}
\definecolor{rowband}{HTML}{F4F6F8}   %

\title{\ours: A Multi-Stream World-Action Model with Compute Flexibility}

\ifpreprint
  \newcommand{\authline}[1]{\makebox[\dimexpr\textwidth-4\tabcolsep\relax][c]{#1}}
\else
  \newcommand{\authline}[1]{#1}
\fi

\author{
  \authline{\bf Ge Yan$^{*\lozenge}$, Jinghao Liu$^{*\lozenge}$, Yuzhi Fan$^{*\lozenge}$, Lei Cai$^{\lozenge}$, Minwen Liao$^{\lozenge}$,} \\
  \authline{\bf Jesse Zhang$^{\dagger\lozenge}$, Dieter Fox$^{\dagger\lozenge \ddagger}$} \\
  \authline{$^{\lozenge}$University of Washington \quad
            $^{\ddagger}$Allen Institute for AI} \\
  \authline{\small $^{*}$Equal contribution. \quad $^{\dagger}$Equal advising.} \\
  \authline{\small\textcolor{urlred}{\url{https://flex-pi.github.io/}}} \\
}

\begin{document}
\maketitle

\ifpreprint\lhead{Preprint.}\fi

\etocdepthtag.toc{mainmatter}

\begin{abstract}
    World-action models (WAMs) predict the future to act better, but nearly all of them predict only RGB latents---trained purely for pixel reconstruction, with no explicit signal for the 3D geometry or object semantics manipulation needs.
    We find a surprising free lunch: the same frozen video-generation VAE that encodes RGB also encodes 3D pointmaps almost losslessly, with no pointmap-specific training at all.
    This lets us supervise \ours{}, a 6B-parameter WAM, on 3D geometry and object-centric DINO semantics alongside RGB---at no cost in new sensors, new pre-training, or inference latency.
    Every visual signal is projected into this shared latent space and denoised jointly with actions inside a Mixture-of-Transformers backbone; per-stream dropout with \emph{cross-modality forcing} then lets a single trained checkpoint run on any subset of these streams, from a fast action-only mode to full joint generation.
    The result is a policy that is exceptionally demonstration-efficient and generalizes well, beating the strongest baselines by $\mathbf{2}$-$\mathbf{6\times}$ on dexterous, precise, real-world bimanual manipulation tasks both in and out of distribution---all while running \emph{faster} than $\pi_{0.5}$.
\end{abstract}

\section{Introduction}
\label{sec:intro}
Generalist robot policies that jointly predict future observations and actions---\emph{world-action models} (WAMs)---have recently emerged as a strong alternative to vision-language-action models (VLAs).
WAMs owe their demonstration efficiency and generalization to two factors: they learn robust representations via \emph{joint prediction} of actions and future observations during training~\citep{zhu2025uwm, ye2026dreamzero}, and they inherit strong spatiotemporal priors from video-generation backbones pre-trained on large-scale video data~\citep{kim2026cosmospolicy, yuan2026fast, nvidia2026cosmos3omnimodalworld}.
We build on both factors to train WAM policies that are substantially more demonstration-efficient and generalizable, while remaining fast enough to deploy.

Although predicting future visual observations is central to WAM training, current generalist WAMs almost exclusively predict future RGB image latents from video generation model encoders~\citep{ye2026gigaworld, ye2026dreamzero, yuan2026fast}.
While effective, these latents, trained via pixel \emph{reconstruction}, mainly capture appearance details and are not aligned with the 3D structure or object semantics needed for robot manipulation.
Supervising geometry and object semantics \emph{directly} would supply both, giving the WAM a stronger joint prediction training signal, but at a steep price: additional sensor modalities, training new priors, or sacrificing inference speed. 
This paper asks: how can we amplify the strengths of WAMs---joint prediction and strong visual priors---without these sacrifices?

To address these questions, we introduce \textbf{\ours{}}, a 6B parameter WAM that is highly demonstration-efficient and generalizable through training to predict not only future RGB observations, but also future 3D pointmaps and object-level semantics.
\ours{} incurs none of the three costs above.
Pointmaps and object semantics are both derived from the same RGB image---via Depth Anything 3~\citep{lin2025depth} and DINOv3~\citep{dinov3} respectively---so no sensor modality beyond RGB is required.
Both encoders are off-the-shelf, so no new visual prior needs to be trained.
And because any visual modality can be dropped at deployment while the model still benefits from the extra training supervision, no inference latency is added.

\begin{figure}[t]
    \centering
    \includegraphics[width=\linewidth]{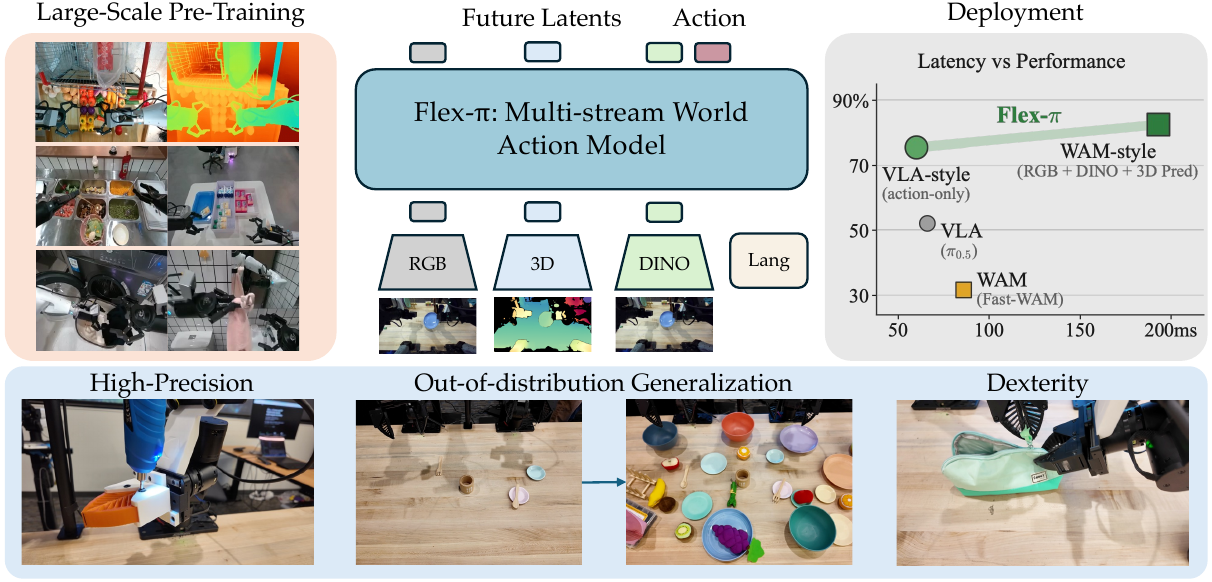}
    \caption{\ours{} is a multi-stream world-action model which can take in RGB, 3D, and DINO visual features to jointly generate both latent future visual features and actions. After training, it supports flexible inference modes that let end users trade off latency and performance.}
    \label{fig:teaser}
\end{figure}

Specifically, \ours{} uses a single, frozen VAE from a pre-trained video generation model~\citep{wan2025} to encode both RGB images and 3D pointmaps into the same latent space---the VAE directly reconstructs pointmaps despite being trained only on RGB pixels---along with DINOv3~\citep{dinov3} to construct object-centric DINO features.
\ours{} embeds every visual modality into a single, shared latent space of token \emph{streams}, one for each modality.
It then routes each stream through a Mixture-of-Transformers~\citep{liang2025mixtureoftransformers} backbone.
Because actions are generated jointly with each future visual stream, the policy inherits rich priors from internet-scale pre-training and learns a stronger internal representation for action generation.
Finally, \ours{} randomly drops out visual input streams during training and applies \emph{cross-modality forcing}---generating each future stream whether or not it was observed as input---so that a single trained checkpoint has the flexibility to operate on any subset of available visual inputs and outputs at inference.
This lets practitioners choose their own point on the speed--performance frontier at deployment time, rather than fixing it during training (\Cref{fig:teaser}).

Our ablations demonstrate that both additional visual streams significantly improve performance, \emph{even when not predicted at inference time}. 
Initialized from a video generation model~\citep{wan2025} and pre-trained on AGIBOT World~\citep{contributors2024agibotworldrepo}, \ours{} shows strong \emph{demonstration efficiency}, \emph{generalization}, and \emph{deployment flexibility}:
In RoboTwin, \ours{} exceeds the strongest WAM baseline by $\mathbf{1.9\times}$ with limited demonstrations and continues to outperform when given full data coverage.
In LIBERO~\citep{liu2023libero}, one \ours{} checkpoint outperforms all existing VLA or WAM methods with up to $99.2\%$ overall success rate while generalizing well to LIBERO-Plus~\citep{fei25libero-plus}.
Finally, in dexterous, precise, real-world tasks on a bimanual YAM, \ours{} outperforms baselines ($\pi_{0.5}$, ManiFlow, Fast-WAM) by $\mathbf{2}$-$\mathbf{6\times}$ in success rates both in and out of distribution.
In fact, when only generating actions, \ours{} achieves lower inference latency than $\pi_{0.5}$ while outperforming all real-world baselines; predicting all visual streams further boosts performance.
Overall, \ours{} shows that a WAM can gain semantic and geometric grounding essentially for free---no new sensors, no new pre-training, no slower inference---resulting in better policy performance while remaining fast enough to deploy.

\section{Related Work}
\label{sec:related}

\textbf{Generalist Manipulation Policies.}
A prominent line of work builds generalist manipulation policies on top of pre-trained vision-language models, %
demonstrating that web-scale priors transfer to robot control with improved generalization over specialist policies~\citep{brohan2023rt2visionlanguageactionmodelstransfer, black2024pi0visionlanguageactionflowmodel, niu2024llarva, intelligence2025pi05, bjorck2025gr00t, hamster2025, goyal2025vla0, geminiroboticsteam2025geminiroboticsbringingai, Chen25-ecot-lite, lee2025molmoact, yan2025maniflow, fang2026molmoact2, zha2026lap, chen2026steerable, kim2026rldx, lingbotvla2, barreiros2026lbm, galaxea2026g05}. 
During action fine-tuning, they predict actions without modeling future observations.

A second line of work predicts how visual inputs change, either by planning in image space or by regularizing the policy through a joint world--action objective.
Early approaches cast control as text-conditioned video generation~\citep{du2023unipi} or pre-train video generators on web data before fine-tuning for manipulation~\citep{gr1, gr2, robodreamer}.
More recent work scales this recipe~\citep{dreamgen, huang2025adapower, bi2025motus, gao2026dreamdojo, yin2026playworld, wang2026interactive, guo2026ctrlworld, zhang2026native}. 
Specifically, joint world--action models such as UWM~\citep{zhu2025uwm} and DreamZero~\citep{ye2026dreamzero} perform action and video generation simultaneously, thereby learning better representations than action generation alone.
For compute efficiency, most world-model-based policies predict in a latent representation space~\citep{zhou2024dino, maes2026leworldmodel, hafner2020mastering, hafner2023mastering, hansen2024tdmpc, Hansen2025Newt, zhu2025uwm}. 
Closest to ours, recent, generalist WAM policies are initialized from pre-trained video-generation models with latent RGB prediction objectives~\citep{ye2026dreamzero, ye2026gigaworld, lingbot-va2026, zhang2026native, yuan2026fast}. 
However, rather than predicting a single RGB latent stream, \ours{} co-denoises latents for RGB, DINO features representing object semantics, and 3D pointmaps, extending WAM training supervision to additionally focus on geometry and semantics.

\textbf{Multimodal Policy Learning.}
A growing body of work grounds manipulation using additional inputs, e.g., by providing explicit 3D inputs~\citep{shridhar2022peract, zhugroot2023, shi2023robocook, goyal2024rvt, Ze2024DP3, yan2025dnact, li2025integrating, 3dvla, singh2025og, qu2025spatialvla, yan2025maniflow} or learning 3D world models~\citep{peripcwm2024, huang2025particleformer, huang2026pointworld, duisterhof2026modality}.
In contrast, \ours{} projects 3D features directly into a shared latent space with DINO and RGB features, using a pre-trained video world model's VAE encoder to encode 3D pointmaps of the same shape as their RGB image counterparts.
Furthermore, \ours{} works even without 3D input at inference time.

Other works demonstrate that masking additional input modalities during training, e.g., through diffusion noising, enables policies to generalize better~\citep{block2023provable, hong2026tmrl} or vary inference-time input modalities~\citep{skand2024simple, huang2025multimodal}.
We drop both visual inputs and outputs via attention masking, training \ours{} to still predict visual modalities that are absent in the input (cross-modality forcing, \Cref{sec:method:dropout}), which we find to \emph{improve policy performance (\Cref{fig:forcing_ablation})} by forcing the model to internalize each visual representation from the others.

\section{\ours{}: A Compute Flexible, Multi-Stream WAM} %
\begin{figure}[t]
    \centering
    \includegraphics[width=\linewidth]{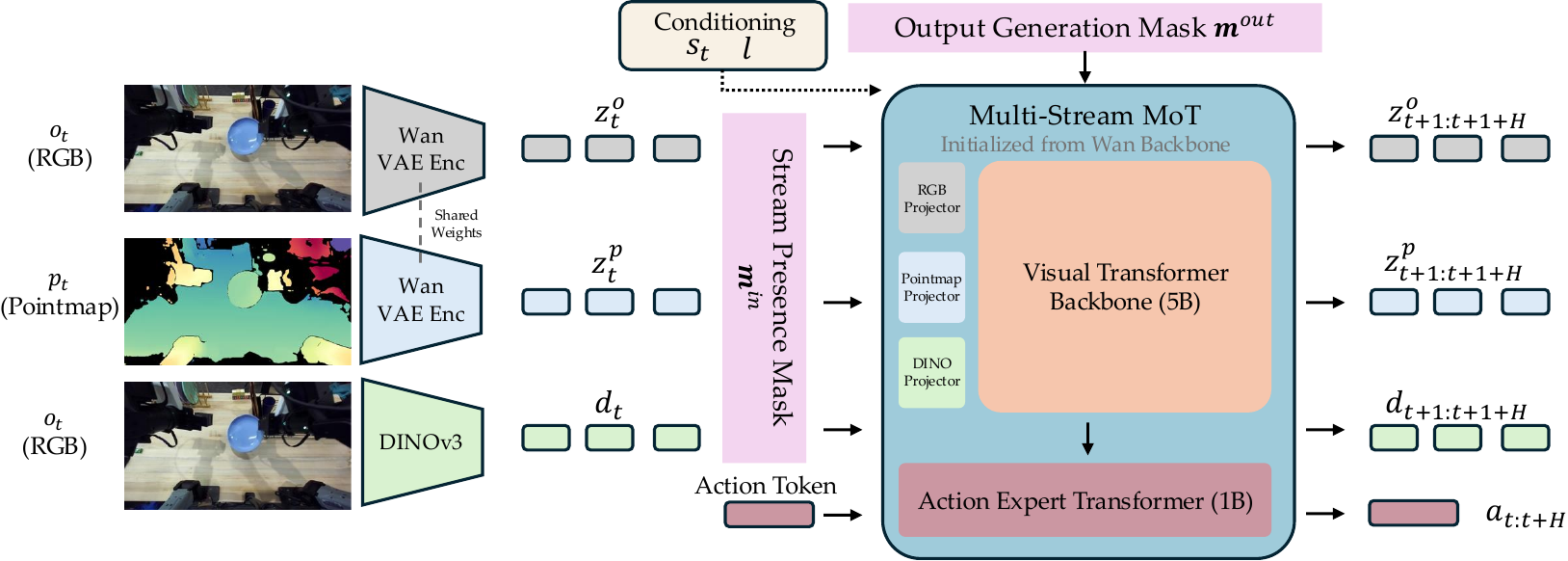}
    \caption{\textbf{\ours{} architecture.}
At time $t$, RGB $o_t$ and 3D pointmaps $p_t$ are encoded by a pre-trained Wan-2.2 VAE into latent token streams $z^o_t, z^p_t$; DINOv3 produces semantic tokens $d_t$.
A stream presence mask $\inputmask$ selects which visual streams are attended to in a shared Visual Transformer Backbone.
A smaller \emph{Action Expert} cross-attends to visual streams to produce an action chunk $a_{t:t+H}$.
Conditioning ($s_t$, $l$) is shared across streams, and an output mask $\outputmask$ sets which future visual streams $\{z^o, z^p, d\}_{t+1}^{t+1+H}$ are mutually visible when jointly generated with the actions.}
\label{fig:method}
\end{figure}

\label{sec:method}
We instantiate \textbf{\ours} as a world-action model supervised on future 3D geometry and object semantics in addition to RGB, without incurring any of the three costs raised in \Cref{sec:intro}: additional sensor inputs, separately trained visual priors, or added inference latency.
A single frozen VAE from a pre-trained video generation model encodes both RGB images and 3D pointmaps, and a frozen DINOv3 encoder supplies object-level semantic tokens; each visual input modality forms a token \emph{stream} fused by a Mixture-of-Transformers backbone initialized from the same video model
(\Cref{sec:method:streams} and \Cref{fig:method}).
We mask out both inputs and outputs during training, yielding a \emph{single checkpoint} that supports any combination of input and output streams at inference (\Cref{sec:method:dropout}).
We detail the training objective, pre-training/fine-tuning protocol, and inference procedure in \Cref{sec:method:objective}.

\textbf{Problem Statement.}
Our goal is to train a policy to predict actions $a_t$ given image observations $o_t$, proprioception $s_t$, and language instruction $l$. %
Additionally, we assume access to DINO features $\dino_t$ from a DINO-v3~\citep{dinov3} encoder and 3D information in the form of pointmaps $\point_t \in \mathbb{R}^{H, W, 3}$ (coming from Depth Anything 3~\citep{lin2025depth} applied to $o$).\footnote{For readibility, we refer to a single $o_t$, $\point_t$, and $\dino_t$, but each observation can consist of multiple tokens.} %
We train a flexible \emph{world action model} (WAM) $\pi_\theta(a_{t}, o_{t+1}, \point_{t+1}, \dino_{t+1} \mid o_t, \point_t, \dino_t, s_t, l)$ which can also generate any subset of future visual input modalities and can minimally take in just a single visual input, e.g., $\pi_\theta(a_t \mid o_t, s_t, l)$.
Note that $a_t$ corresponds to an action chunk $a_{t:t+H}$ (same with output observations), but we drop chunk notation throughout most of the paper for readibility.

\textbf{Flow Matching for Video Generation Models.}
State-of-the-art video generation models~\citep{HaCohen2024LTXVideo, kong2024hunyuanvideo, wan2025, nvidia2025cosmospredict2, nvidia2026cosmos3omnimodalworld} often leverage an image
encoder $\enc$ and decoder $\dec$, e.g., a set of U-Nets~\citep{unet2015}
trained as part of a variational auto-encoder (VAE)~\citep{Kingma2013AutoEncodingVB},
to encode images $o_{1:t}$ into a latent space $z_{1:t} = \enc(o_{1:t})$,
predict future latent images $z_{t+1}$,
and then decode them back into 
reconstructed images $o_{t+1} = \dec(z_{t+1})$. Latent prediction of the ground truth next latent is performed
with a flow matching~\citep{lipman2023flow,liu2023flow} diffusion transormer~\citep{peebles2022DiT} model
$\latentmodel(z_{t+1}^{\tau} \mid \tau,  l, z_{\le t})$ trained on the linear path
$z^{\tau}_{t+1} = \tau z_{t+1} + (1-\tau)\epsilon$ with $\epsilon \sim \mathcal{N}(0, I)$
and flow timestep $\tau \in [0, 1]$, regressing the flow path's constant velocity $z_{t+1} - \epsilon$:
\begin{equation}
    \mathcal{L}_{\text{FM}}(z_{t+1}) =
    \mathbb{E}_{z, \epsilon, \tau}
    \big\| \latentmodel(\cdot \mid \tau, l, z_{1:t}) - (z_{t+1} - \epsilon) \big\|_2^2.
    \label{eq:fm-loss}
\end{equation}
At inference, $z_{t+1}$ is generated by integrating $\latentmodel$ from $\tau{=}0$ to $\tau{=}1$ with $K$ Euler steps and decoded
via $\dec$. Crucially, $(\enc, \dec)$ defines a latent space for \emph{any}
image-shaped tensor; we use it for both images $o_t$ and pointmaps $\point_t$.
Note that 
sampling from $\pi$ corresponds to integrating $\latentmodel$ along the flow ODE over active output streams.

\subsection{Input Visual Streams and Model Backbone}
\label{sec:method:streams}
\ours{} operates over three complementary visual token streams, each contributing visual, spatial, or object-level semantic information.
RGB images $o_t$ and 3D pointmaps $\point_t$ add appearance context and explicit spatial geometry respectively, and are both encoded by the \emph{same frozen VAE} from a pretrained video generation model so that both streams directly inherit spatiotemporal priors from large-scale video pretraining.
DINO features $\dino_t$, computed by a frozen DINOv3~\citep{dinov3} encoder, help ground the other features with object semantics.

\begin{wrapfigure}[8]{R}{0.45\textwidth}
    \vspace{-1em}
    \includegraphics[width=\linewidth]{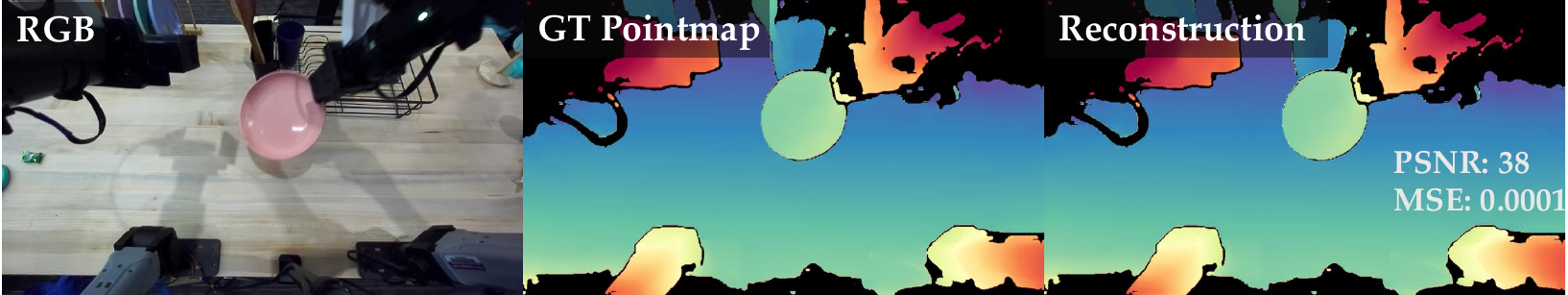}
    \caption{\textbf{Wan-VAE pointmap reconstruction} closely matches the ground truth, despite only being trained on RGB images.}
    \label{fig:3d_inputs}
\end{wrapfigure}
\textbf{Visual Input Encoding.} We use the frozen encoder and decoder from the VAE of the \texttt{Wan-2.2-5B}~\citep{wan2025} video generation model.
Surprisingly, we find that directly encoding and then decoding pointmaps with this frozen VAE---trained only on images---yields very accurate reconstructions, making its latent space immediately useful for encoding 3D information while preserving scene structure (\Cref{fig:3d_inputs}).
For DINO we keep all patch tokens and losslessly \emph{fold} each $2{\times}2$ neighborhood of patches into a single token ($768 \to 3072$ dim., cf. \texttt{PixelUnshuffle}~\citep{shi2016pixleshuffle}), cutting the tokens the model generates by $4\times$ without discarding spatial detail (\Cref{sec:appdx:adapters}).

\textbf{Global Conditioning.}
Text instructions $l$ are processed by a frozen \texttt{umT5} encoder~\citep{chung2023unimax} from \texttt{Wan-2.2} and combined with $s_t$ (encoding details in \Cref{sec:appdx:proprioception}) to serve as global conditioning vectors for the latent prediction model $\latentmodel$ ({\color{conditioning}\rule[-0.05em]{0.8em}{0.8em}} in \Cref{fig:method}).
\ours{}'s inputs are:
\begin{equation}
\begin{array}{l@{\;\;}l@{\quad}l}
    \text{RGB latent tokens}      & z^o_t    = \enc(o_t)      & \triangleright \, \text{\algcomment{Appearance; inherits video pretraining priors}} \\[2pt]
    \text{Pointmap latent tokens} & z^\point_t = \enc(p_t)    & \triangleright \, \text{\algcomment{3D geometry; shared video VAE latent space}}    \\[2pt]
    \text{DINO tokens}   & d_t      = \dinoenc(o_t)  & \triangleright \, \text{\algcomment{Object/semantic grounding}}                     \\[2pt]
    \text{Proprioception}  & s_t                       & \triangleright \, \text{\algcomment{Robot joint states (global conditioning for $\latentmodel$)}}  \\[2pt]
    \text{Language}        & l                         & \triangleright \, \text{\algcomment{Task specification (global conditioning for $\latentmodel$)}}
\end{array}
\label{eq:inputs}
\end{equation}

\textbf{Model Backbone.}
Given the inputs in \Cref{eq:inputs}, we train a latent DiT model $\latentmodel(a_{t+1}, z_{t+1}^o, z^\point_{t+1}, d_{t+1} \mid s_t, l)$ which predicts both future actions and RGB, pointmap, and DINO latent tokens. Jointly predicting future images has been shown to improve action prediction~\citep{zhu2025uwm, ye2026dreamzero, dyna2026dyna2}; we show the same holds for both pointmaps and DINO features in our experiments.
In order to reduce training and inference time while fusing all input and output modalities, we employ a \emph{mixture-of-transformers} (MoT)~\citep{liang2025mixtureoftransformers} model $\latentmodel$ ({\color{mot}\rule[-0.05em]{0.8em}{0.8em}} in \Cref{fig:method}) that processes all of the visual modality tokens $z^o, z^p, d_t$ with the same transformer backbone (5B parameters, initialized directly from the pre-trained \texttt{Wan-2.2-5B}) and independent, per-modality feedforward projection networks, but contains separate key, query, value, feedforward, and normalization parameters for action generation ($\sim$1B parameters; we reduce the hidden dimension).
Cross-stream attention is applied only in the middle $16$ of the $30$ MoT blocks, so early encoding and late decoding stay stream-specific while cross-modal fusion happens in the trunk.
All outputs $a, z^o, z^\point, d$ are finally decoded by stream-specific feedforward prediction heads trained with the flow matching loss (\Cref{eq:fm-loss}).
Because the action expert is narrower than the backbone, we initialize it from \texttt{Wan-2.2} by \emph{resampling} rather than copying~\citep{yuan2026fast}; \Cref{sec:appdx:adapters} details this procedure and the per-stream adapters that map each modality into and out of the shared trunk.

\textbf{Causal Joint Generation.}
All outputs $z^o_{t+1}, z^\point_{t+1}, d_{t+1}, a_t$ are \emph{jointly generated}: action tokens attend to current observation tokens $z^o_t, z^\point_t, d_t$ and cross-attend to whichever visual output tokens are active, so action generation benefits from the model's evolving representation of the future~\citep{zhu2025uwm, ye2026dreamzero}.
Attention is one-way---no visual token ever attends to the action stream---see full attention rules in \Cref{sec:appdx:regimes}.

\subsection{Flexible Training via Visual Stream Dropout and Cross-Modality Forcing}
\label{sec:method:dropout}
A single \ours{} checkpoint supports any combination of available input and output streams at inference---from action-only to full joint generation, even with just 1 visual input---without retraining. We achieve this with visual stream dropout and cross-modality forcing, detailed below.

\textbf{Visual Stream Dropout.}
We construct two independent, per-sample binary masks over the three latent visual streams, RGB ($z^o$), DINO ($d$), and pointmap ($z^\point$): the \emph{input presence mask} 
$\inputmask \in \{0,1\}^3$ 
selects which streams are provided as conditioning at time $t$, and the \emph{output attention mask} 
$\outputmask \in \{0,1\}^3$ 
({\color{masks}\rule[-0.05em]{0.8em}{0.8em}} in \Cref{fig:method}) governs the \emph{joint-generation attention} among the future tokens: both what the action tokens cross-attend to and how the future visual tokens attend to one another.
Every visual input is dropped independently with probability $0.5$, subject to at least one visual stream remaining observed (\Cref{sec:appdx:regimes}).
Crucially, $\outputmask$ is not a loss mask: it selects which futures the action tokens read, but every future stream is always denoised and incurs its flow-matching loss (\Cref{eq:total-loss}).
Those futures attend to one another and to whichever streams are observed at time $t$, so actions are conditioned on a jointly generated future rather than three independent ones.

\begin{figure}[t]
\centering

\definecolor{strV}{HTML}{6F7479}\definecolor{strVf}{HTML}{DEE0E1}
\definecolor{strD}{HTML}{2E7D4F}\definecolor{strDf}{HTML}{D9F1D0}
\definecolor{strP}{HTML}{4E82AE}\definecolor{strPf}{HTML}{DDE9F4}
\definecolor{strA}{HTML}{C4685C}\definecolor{strAf}{HTML}{F7D8D2}
\definecolor{maskpink}{HTML}{E59EDD}
\definecolor{edgedark}{HTML}{3A3F45}
\definecolor{cutgrey}{HTML}{C8CDD2}
\definecolor{obsgrey}{HTML}{8A9299}
\definecolor{starpink}{HTML}{B02E8A}

\newcommand{\rgfam}{\fontfamily{ppl}\selectfont}
\newcommand{\rgtxt}{\rgfam\fontsize{7.5}{8.6}\selectfont}
\newcommand{\rgsub}{\rgfam\fontsize{6.6}{7.6}\selectfont}
\newcommand{\rgglyph}{\rgfam\fontsize{6.2}{7}\selectfont\vphantom{dp}}
\newcommand{\rghead}{\rgfam\fontsize{8}{9}\selectfont}

\newcommand{\rgchipon}[4]{%
  \node[draw=#3,fill=#3f,minimum width=4.2mm,
        minimum height=4.2mm,inner sep=0pt,line width=0.45pt]
        at (#1,#2) {\rgglyph #4};}
\newcommand{\rgchipoff}[4]{%
  \node[draw=#3,fill=white,densely dashed,
        minimum width=4.2mm,minimum height=4.2mm,inner sep=0pt,line width=0.6pt]
        at (#1,#2) {\rgglyph #4};}
\newcommand{\rgbit}[5]{\ifnum#1=1 \rgchipon{#2}{#3}{#4}{#5}\else\rgchipoff{#2}{#3}{#4}{#5}\fi}

\newcommand{\rgcut}[2]{%
  \draw[maskpink,line width=0.85pt]
    (#1-0.75,#2-0.75)--(#1+0.75,#2+0.75) (#1-0.75,#2+0.75)--(#1+0.75,#2-0.75);}

\newcommand{\rgff}[8]{%
  \ifnum#1=#2 \draw[edgedark,line width=0.95pt] (#3,#4)--(#5,#6);
  \else
    \draw[cutgrey,line width=0.75pt,densely dashed] (#3,#4)--(#5,#6);
    \rgcut{#7}{#8}%
  \fi}

\newcommand{\rgnode}[5]{%
  \ifnum#1=1
    \node[circle,draw=#4,fill=#4f,inner sep=0pt,minimum size=4.4mm,line width=0.45pt]
      at (#2,#3) {\rgglyph #5};
  \else
    \node[circle,draw=#4,fill=white,densely dashed,inner sep=0pt,minimum size=4.4mm,
          line width=0.55pt] at (#2,#3) {\rgglyph #5};
  \fi}

\newcommand{\rgmicro}[8]{%
\begin{scope}[shift={(#7,#8)}]
  \draw[obsgrey,line width=0.7pt,densely dotted] (4.5,0) -- (19,4.0);
  \draw[obsgrey,line width=0.7pt,densely dotted] (4.5,0) -- (14,-3.4);
  \draw[obsgrey,line width=0.7pt,densely dotted]
        (4.5,0) .. controls (13,-0.5) and (19,-1.6) .. (25,-3.4);
  \draw[obsgrey,line width=0.7pt,densely dotted] (4.5,0) -- (38.2,0);
  \rgff{#4}{#5}{19}{4.0}{14}{-3.4}{16.5}{0.3}
  \rgff{#4}{#6}{19}{4.0}{25}{-3.4}{22}{0.3}
  \rgff{#5}{#6}{14}{-3.4}{25}{-3.4}{19.5}{-3.4}
  \ifnum#4=1 \draw[edgedark,line width=0.95pt,-{Latex[length=1.5mm,width=1.2mm]}]
        (19,4.0)--(38.2,1.7);
  \else \draw[cutgrey,line width=0.75pt,densely dashed] (19,4.0)--(38.2,1.7);
        \rgcut{28.6}{2.85}\fi
  \ifnum#5=1 \draw[edgedark,line width=0.95pt,-{Latex[length=1.5mm,width=1.2mm]}]
        (14,-3.4) .. controls (23,-8.2) and (33,-4.6) .. (38.2,-1.7);
  \else \draw[cutgrey,line width=0.75pt,densely dashed]
        (14,-3.4) .. controls (23,-8.2) and (33,-4.6) .. (38.2,-1.7);
        \rgcut{24.9}{-5.7}\fi
  \ifnum#6=1 \draw[edgedark,line width=0.95pt,-{Latex[length=1.5mm,width=1.2mm]}]
        (25,-3.4)--(38.2,0);
  \else \draw[cutgrey,line width=0.75pt,densely dashed] (25,-3.4)--(38.2,0);
        \rgcut{31.6}{-1.7}\fi
  \draw[draw=black!30,fill=white,line width=0.4pt]
        (0.5,-5.0) rectangle (4.5,5.0);
  \ifnum#1=1 \draw[draw=strV,fill=strVf,line width=0.4pt] (1.2,1.4) rectangle (3.8,4.6);
  \else \draw[strV,densely dashed,line width=0.5pt] (1.2,1.4) rectangle (3.8,4.6);\fi
  \ifnum#2=1 \draw[draw=strD,fill=strDf,line width=0.4pt] (1.2,-1.6) rectangle (3.8,1.6);
  \else \draw[strD,densely dashed,line width=0.5pt] (1.2,-1.6) rectangle (3.8,1.6);\fi
  \ifnum#3=1 \draw[draw=strP,fill=strPf,line width=0.4pt] (1.2,-4.6) rectangle (3.8,-1.4);
  \else \draw[strP,densely dashed,line width=0.5pt] (1.2,-4.6) rectangle (3.8,-1.4);\fi
  \rgnode{#4}{19}{4.0}{strV}{$z^o$}
  \rgnode{#5}{14}{-3.4}{strD}{$d$}
  \rgnode{#6}{25}{-3.4}{strP}{$z^p$}
  \node[draw=strA,fill=strAf,minimum width=3.6mm,
        minimum height=6.2mm,inner sep=0pt,line width=0.45pt] at (40,0)
        {\rgglyph $a$};
\end{scope}}

\begin{tikzpicture}[x=1mm,y=1mm]

\fill[maskpink!18] (0,5.5) rectangle (139.0,19);

\node[text=black!70,inner sep=0pt] at (12,64.0) {\rghead Training regime};
\node[text=black!70,inner sep=0pt] at (35.6,64.0) {\rghead $\inputmask$};
\node[text=black!70,inner sep=0pt] at (54.3,64.0) {\rghead $\outputmask$};
\node[text=black!70,inner sep=0pt] at (70.0,64.0) {\rghead obs.\ at $t$};
\node[text=black!70,inner sep=0pt] at (94.0,64.0) {\rghead Attention among futures};
\node[font=\rghead,text=black!70,align=center,inner sep=0pt] at (128.4,62.4)
  {Computed\\at inference};

\node[font=\rgtxt,text=black!70,align=center,inner sep=0pt] at (12,52.75)
  {full joint\\generation};
\rgbit{1}{30.6}{52.75}{strV}{$z^o$} \rgbit{1}{35.6}{52.75}{strD}{$d$} \rgbit{1}{40.6}{52.75}{strP}{$z^p$}
\rgbit{1}{49.3}{52.75}{strV}{$z^o$} \rgbit{1}{54.3}{52.75}{strD}{$d$} \rgbit{1}{59.3}{52.75}{strP}{$z^p$}
\rgmicro{1}{1}{1}{1}{1}{1}{67.5}{52.75}
\rgbit{1}{120.9}{52.75}{strV}{$z^o$} \rgbit{1}{125.9}{52.75}{strD}{$d$}
\rgbit{1}{130.9}{52.75}{strP}{$z^p$} \rgchipon{135.9}{52.75}{strA}{$a$}

\node[font=\rgtxt,text=black!70,align=center,inner sep=0pt] at (12,39.25)
  {action-only\\generation};
\rgbit{1}{30.6}{39.25}{strV}{$z^o$} \rgbit{1}{35.6}{39.25}{strD}{$d$} \rgbit{1}{40.6}{39.25}{strP}{$z^p$}
\rgbit{0}{49.3}{39.25}{strV}{$z^o$} \rgbit{0}{54.3}{39.25}{strD}{$d$} \rgbit{0}{59.3}{39.25}{strP}{$z^p$}
\rgmicro{1}{1}{1}{0}{0}{0}{67.5}{39.25}
\rgbit{0}{120.9}{39.25}{strV}{$z^o$} \rgbit{0}{125.9}{39.25}{strD}{$d$}
\rgbit{0}{130.9}{39.25}{strP}{$z^p$} \rgchipon{135.9}{39.25}{strA}{$a$}

\node[font=\rgtxt,text=black!70,align=center,inner sep=0pt] at (12,25.75)
  {e.g.\ 3D $+$ action\\joint generation};
\rgbit{1}{30.6}{25.75}{strV}{$z^o$} \rgbit{1}{35.6}{25.75}{strD}{$d$} \rgbit{1}{40.6}{25.75}{strP}{$z^p$}
\rgbit{0}{49.3}{25.75}{strV}{$z^o$} \rgbit{0}{54.3}{25.75}{strD}{$d$} \rgbit{1}{59.3}{25.75}{strP}{$z^p$}
\rgmicro{1}{1}{1}{0}{0}{1}{67.5}{25.75}
\rgbit{0}{120.9}{25.75}{strV}{$z^o$} \rgbit{0}{125.9}{25.75}{strD}{$d$}
\rgbit{1}{130.9}{25.75}{strP}{$z^p$} \rgchipon{135.9}{25.75}{strA}{$a$}

\node[font=\rgtxt\bfseries,text=black!70,align=center,inner sep=0pt] at (12,12.25)
  {cross-modality\\forcing};
\rgbit{1}{30.6}{12.25}{strV}{$z^o$} \rgbit{1}{35.6}{12.25}{strD}{$d$} \rgbit{0}{40.6}{12.25}{strP}{$z^p$}
\rgbit{1}{49.3}{12.25}{strV}{$z^o$} \rgbit{0}{54.3}{12.25}{strD}{$d$} \rgbit{1}{59.3}{12.25}{strP}{$z^p$}
\rgmicro{1}{1}{0}{1}{0}{1}{67.5}{12.25}
\rgbit{1}{120.9}{12.25}{strV}{$z^o$} \rgbit{0}{125.9}{12.25}{strD}{$d$}
\rgbit{1}{130.9}{12.25}{strP}{$z^p$} \rgchipon{135.9}{12.25}{strA}{$a$}
\node[text=starpink,inner sep=0pt] at (40.6,15.9) {\fontsize{9}{10}\selectfont $\bigstar$};
\node[text=starpink,inner sep=0pt] at (59.3,15.9) {\fontsize{9}{10}\selectfont $\bigstar$};

\draw[edgedark,line width=0.95pt] (0,2.2)--(4.2,2.2);
\node[anchor=west,text=black!70,inner sep=0pt] at (5.0,2.2) {\rgtxt mutually visible};
\draw[edgedark,line width=0.95pt,-{Latex[length=1.5mm,width=1.2mm]}] (27.5,2.2)--(31.7,2.2);
\node[anchor=west,text=black!70,inner sep=0pt] at (32.5,2.2) {\rgtxt read by action};
\node[circle,draw=black!45,fill=white,densely dashed,inner sep=0pt,minimum size=3.4mm,
      line width=0.55pt] at (54,2.2) {};
\node[anchor=west,text=black!70,inner sep=0pt] at (56.3,2.2) {\rgtxt unread by action};
\draw[obsgrey,line width=0.7pt,densely dotted] (79.5,2.2)--(83.7,2.2);
\node[anchor=west,text=black!70,inner sep=0pt] at (84.5,2.2) {\rgtxt attends to observations};
\rgcut{115.5}{2.2}
\node[anchor=west,text=black!70,inner sep=0pt] at (117.7,2.2) {\rgtxt attention blocked};

\end{tikzpicture}

\caption{\textbf{One checkpoint, any visual input and output.}
Each row is one example training scheme determined by sampling $\inputmask$, selecting which streams are observed
at time $t$, and $\outputmask$, determining what visual outputs are visible for joint prediction.
\textcolor{starpink}{$\bigstar$} shows an example of cross-modality forcing where the pointmap input is not attended to, yet joint generation of other outputs still conditions on generated pointmap futures, encouraging \ours{} to internalize 3D representations. 
}
\label{fig:regimes}
\end{figure}
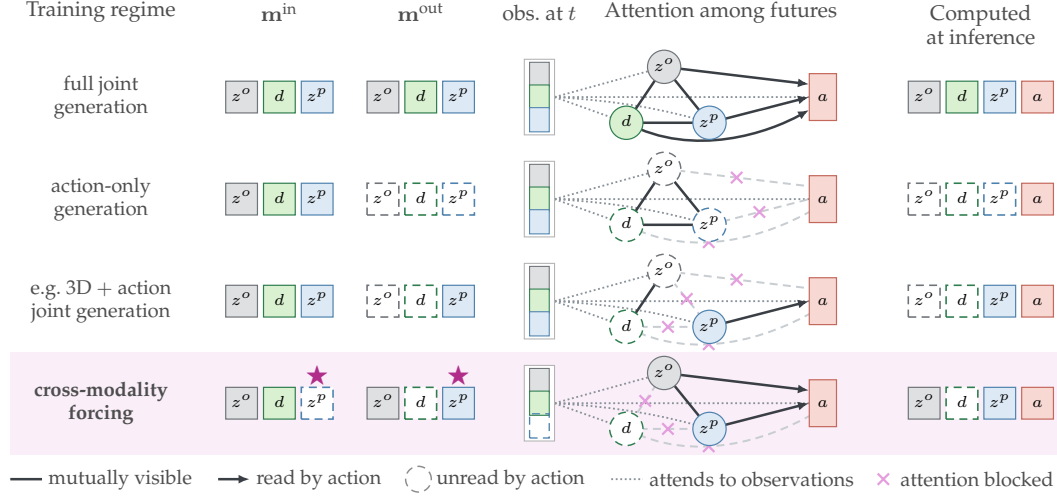

\textbf{Cross-Modality Forcing.}
Because the two masks are drawn independently, a stream dropped from the input is still denoised at the output; the model must synthesize that modality's future from the remaining streams.
We call this \emph{cross-modality forcing} and enable it for all three visual streams, so training covers every mapping from a non-empty subset of $\{z^o, d, z^\point \}$ observed at time $t$ to any subset generated at $t{+}1$: imagining future pointmaps from RGB and DINO with no 3D input, future DINO semantics from RGB and geometry, and so on (see \Cref{fig:regimes}).
Reminiscent of masked language models~\citep{devlin-etal-2019-bert, reimers-2019-sentence-bert} or masked visual autoencoders~\citep{he2022mae}, this constraint forces \ours{} to internalize 3D, RGB, and object-centric representations, thereby aiding policy learning: we show in \Cref{sec:additional visual inputs} that removing it degrades performance.

\subsection{Training Objectives, Data, and Inference}
\label{sec:method:objective}

With the architecture (\Cref{sec:method:streams}) and dropout scheme (\Cref{sec:method:dropout}) in place, training and inference both reduce to running flow matching over whichever streams are active in a given sample.

\textbf{Loss.}
Because every future stream is denoised regardless of $\outputmask$, the training objective sums the per-stream flow-matching loss of \Cref{eq:fm-loss} over the action stream and \emph{all three} visual streams:
\begin{equation}
    \mathcal{L}(\theta) = \lambda_a \mathcal{L}^\text{FM}_a(a_t) \;+\; \sum_{i \,\in\, \{o, d, p\}} \lambda_i\, \mathcal{L}^\text{FM}_i(i_{t+1}),
    \label{eq:total-loss}
\end{equation}
where $\lambda_a, \{\lambda_i\}$ are per-stream loss weights, all set to $1$ in every experiment we report. The output mask $\outputmask$ is only an attention mask, not a loss mask---every stream head still trains on \Cref{eq:total-loss}. 
For DINO we use ``$x$-prediction''~\citep{salimans2022progressive}: the head predicts the clean features $d_{t+1}$ rather than the velocity, since the DINO folding of \Cref{sec:method:streams} raises each token's feature dimension to $3072$, where velocity prediction degrades~\citep{li2025back} (\Cref{sec:appdx:dinoparam}).

\textbf{Pre-training and Fine-tuning.}
We pre-train \ours{} on $\sim500$ hours drawn from $100$ tasks of AGIBOT World-Beta~\citep{contributors2024agibotworldrepo}, a large-scale bimanual manipulation dataset recorded at $30$\,Hz from a head camera and two wrist cameras. 
Pointmaps come from annotating all three views with Depth Anything~3~\citep{lin2025depth}.
See \Cref{sec:appdx:pretraining} for additional pre-training details.
After pre-training, we perform domain-specific fine-tuning for each experiment domain.

\textbf{Inference.}
Given a chosen input/output regime, \ours{} runs $K$ Euler steps of the flow-matching ODE over the active output streams and emits an action chunk of length $H$ (\Cref{sec:appdx:actions}).
\Cref{sec:additional visual inputs} measures what each of these choices costs and buys; \Cref{sec:appdx:inference} describes the training-free deployment stack the reported latencies are measured on, and \Cref{sec:appdx:architecture} gives further architecture details.

\section{Experiments}
\label{sec:experiments}
We design our experiments to answer four questions:
\begin{enumerate}[label=\textbf{(Q\arabic*)}  ]
    \item \label{q1} Is \ourexp{} effective in contact-rich, precise, and long-horizon manipulation on real robots? %
    \item \label{q2} Does \ourexp{} hold that performance under OOD conditions or limited demonstrations?
    \item \label{q3} How does \ourexp{} compare against state-of-the-art VLAs and WAMs across the inference speed-performance spectrum?
    \item \label{q4} How important are \ourexp{}'s additional visual inputs (dino and pointmaps) and their joint generation in policy performance?
\end{enumerate}

\subsection{Experimental Setup}
\label{sec:exp:setup}

\textbf{Benchmarks.} We evaluate on two simulation suites and one real-world platform:
(i)~\textbf{RoboTwin}~\citep{chen2025robotwin} --- a bimanual simulation benchmark with ground-truth 3D, used for the Pareto and data-scaling analyses;
(ii)~\textbf{LIBERO}~\citep{liu2023libero} --- four task suites (Spatial, Object, Goal, Long) testing model fitting capabilities;
(iii)~\textbf{LIBERO-Plus}~\citep{fei25libero-plus} --- a perturbed variant of the LIBERO suites that injects seven categories of visual, spatial, and language shifts to probe generalization;
and (iv)~a real bimanual \textbf{YAM} robot evaluated on five dexterous tasks, with a held-out subset for generalization.

\textbf{Baselines.}
We compare against three representative robot policy baselines on RoboTwin: \pibaseline~\citep{intelligence2025pi05} as the VLA baseline, and \fastwam~\citep{yuan2026fast} and \lingbot~\citep{lingbot-va2026} as representative, similar parameter-count WAM baselines, which unlike \ourexp{}, train only with RGB latent prediction.
We retrain all three at the reduced demonstration budgets of the data-scaling sweep, since no published work reports them, and quote published figures at full data (\Cref{sec:appdx:baselines}).
We additionally report published RoboTwin and LIBERO numbers for a variety of additional WAM and VLA baselines.
In the real world, we again compare against \pibaseline\ and \fastwam\ along with a 3D VLA baseline, \maniflow~\citep{yan2025maniflow}.

\textbf{Training protocol.} All \ourexp{} variants are pre-trained on AGIBOT World~\citep{contributors2024agibotworldrepo} and fine-tuned in each setting. %
Full hyperparameters are in \Cref{sec:appdx:experiments}.

\subsection{\ref{q1}: Real-World Bimanual Manipulation}
\label{sec:exp:real}

\begin{figure}[t]
\centering
\includegraphics[width=\linewidth]{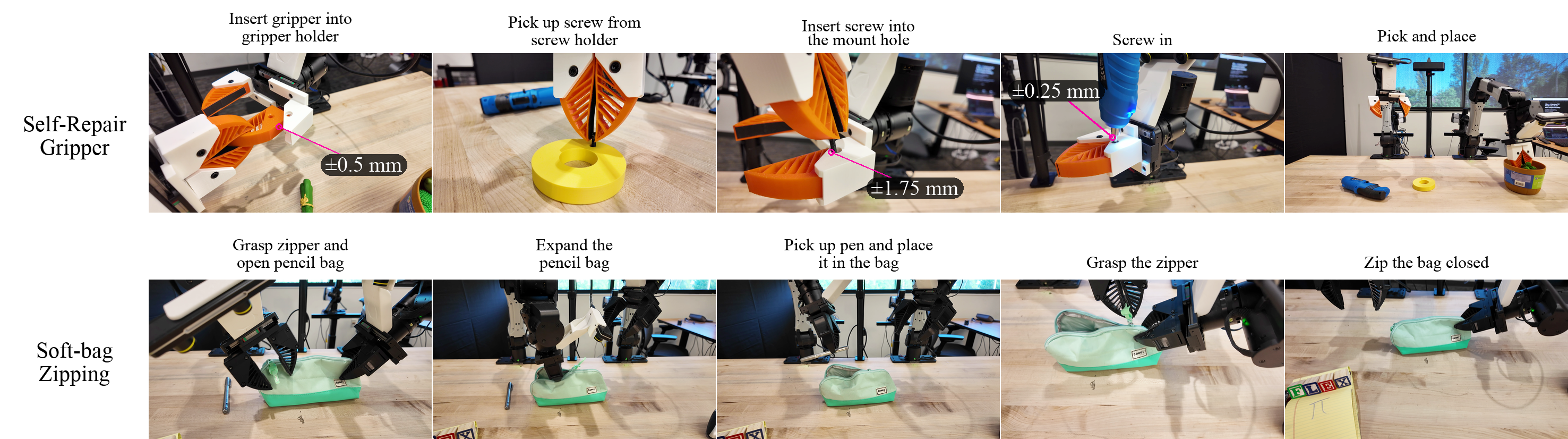}
\caption{\textbf{Illustration of selected evaluation tasks.}
\emph{Self-Repair Gripper} rebuilds the robot's own gripper over eight stages
that must be completed in order, at $\pm 0.25$--$0.5$\,mm of insertion
clearance; \emph{Soft-Bag Zipping} opens a deformable pencil case, places a pen
inside, and zips it shut (\Cref{sec:appdx:addtl_exps}).}
\label{fig:hard_tasks}
\end{figure}

\textbf{Task selection and setup.} We deploy a single pre-trained \ourexp{} checkpoint on a real bimanual YAM robot (\Cref{sec:appdx:yam}) across five tasks that span the difficulty range we can measure. For all methods, we perform per-task fine-tuning.
For fairness, we chose the number of per-task demonstrations to collect by iteratively collecting data and training the smallest baseline method, \maniflow{}, until it achieved reasonable task completion rates.
We also selected tasks for evaluation mode coverage: dexterity, precision, long-horizon, and contact-rich manipulation. 
\textbf{Put Plate on Rack} and \textbf{Sort Utensils} test bimanual coordination under sustained contact, and \textbf{Kitchen Organization} chains four such skills in a single long-horizon episode: placing a plate, inserting a spoon, stacking cups, and a bimanual handover into a rack slot.
Every method trains on all four skills, and we score the two we evaluate under: placing the bowls and the bimanual handover into the rack (\Cref{tab:appdx:rubrics}).
The remaining two are the most difficult (\Cref{fig:hard_tasks}): \textbf{Self-Repair Gripper} has the robot use an electric screwdriver to re-insert and screw in \emph{its own gripper} over eight stages, with insertions having just $\pm 0.25$--$0.5$\,mm of clearance
(\Cref{sec:selfrepair}); \textbf{Soft-Bag Zipping} acts on a deformable pencil case so the target pose changes significantly between every rollout
(\Cref{sec:softbag}).
Every rollout is scored under the partial-credit rubric of
\Cref{app:data-real-world-eval-criteria}. We report \emph{task completion} (the
points earned as a fraction of the maximum attainable) and its standard deviation, along with
\emph{binary success}, the percentage of rollouts which solve the entire task, over $10$--$20$ trials per task (\Cref{sec:q3}).
\setlength{\columnsep}{10pt}

\begin{figure}[t]
    \centering
    \includegraphics[width=\linewidth]{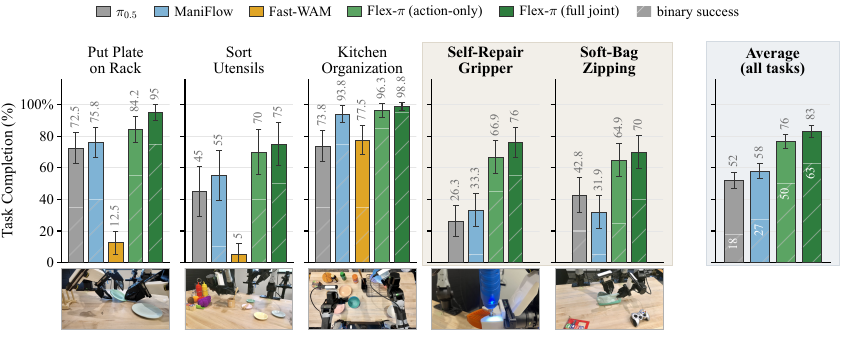}
    \captionsetup{skip=6pt}
    \caption{\textbf{Real-world results.} We report task completion for
    \ourexp{} and three baselines on five bimanual tasks. Each bar is task
    completion; the hatched region at its base is the binary success rate, the
    fraction of rollouts satisfying the entire rubric. The rightmost panel averages
    the five tasks. Both metrics are scored under the rubric of
    \Cref{app:data-real-world-eval-criteria}. \ourexp{} leads on every task, with the
    largest margins on \emph{Self-Repair Gripper} and \emph{Soft-Bag Zipping}, which
    require sub-millimeter precision over a long horizon and dexterous manipulation
    of a deformable object respectively. The action-only variant, the cheapest policy here to run
    (\Cref{fig:real_latency_scatter}), already outperforms every baseline on every
    task.}
    \label{fig:real_main}
    \vspace{-11pt}
\end{figure}

\begin{wrapfigure}[13]{r}{0.33\textwidth}
    \includegraphics[width=\linewidth]{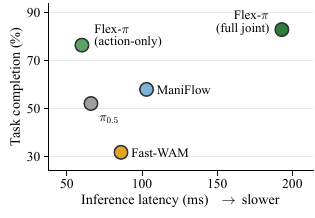}
    \caption{\textbf{Real-world speed--success frontier.} Five-task mean completion against
    measured latency (\Cref{sec:appdx:baseline_latency}).}
    \label{fig:real_latency_scatter}
\end{wrapfigure}

\textbf{\ourexp{} outperforms on every task.} \Cref{fig:real_main} shows both action-only and full joint generation \emph{outperform every baseline} on all five tasks, and the margin grows with task difficulty: with full joint generation, \ourexp{} achieves $+5.0$ points of task completion over the strongest baseline on \emph{Kitchen Organization} but more on difficult tasks: $+42.7$ on \emph{Self-Repair Gripper} and $+27.2$ on \emph{Soft-Bag Zipping}.
Averaged over all tasks, \ours{} achieves $\mathbf{3.5}\times$ higher success rate against \pibaseline{} and $\mathbf{2.3}\times$ against \mbox{\maniflow{}}.
\fastwam{} is not competitive on any of these tasks, hence we do not evaluate it on two most difficult tasks.

\textbf{Inference speed vs task completion.} We compare real-world inference speed vs averaged task completion in \mbox{\Cref{fig:real_latency_scatter}}.
\ourexp{} action-only exceeds the best baseline on all five tasks at $60$\,ms per call, faster than every other baseline, so the multi-stream training objective pays off even when no visual stream is generated at test time (more ablations later in \Cref{sec:additional visual inputs}). 
\Cref{fig:real_perf_latency} gives the same comparison with every measured configuration broken out. Enabling joint generation adds latency but results in higher task completion: it increases success rates by $+13\%$ on average.

\subsection{\ref{q2}: Real World OOD Generalization and Demo Efficiency}
\label{sec:exp:real_gen}

\begin{wrapfigure}{r}{0.28\textwidth}
  \vspace{-\intextsep}
  \centering
  \includegraphics[width=\linewidth]{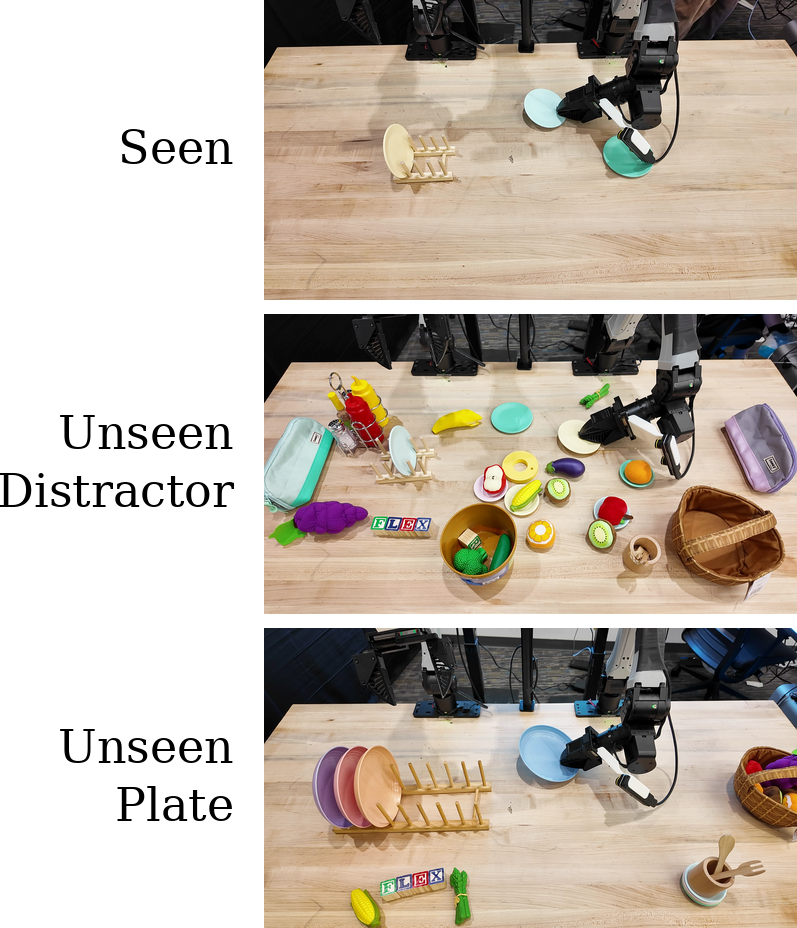}
  \caption{\textbf{Plate on Rack generalization conditions.} }
  \label{fig:plate-rack-conditions}
  \vspace{-\intextsep}
\end{wrapfigure}

\begin{figure}[t]
    \centering
    \includegraphics[width=\linewidth]{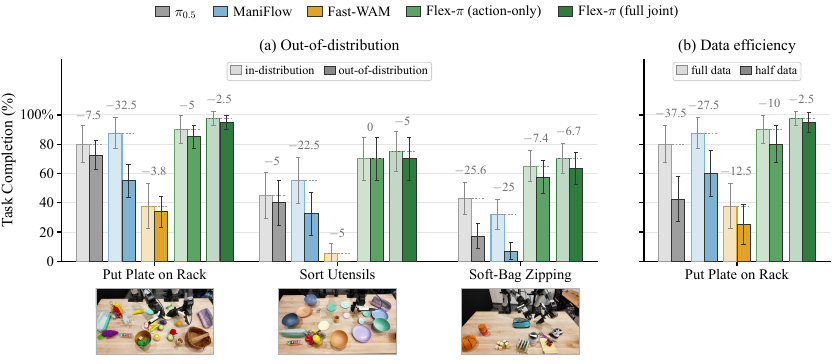}
    \captionsetup{skip=6pt}
    \caption{\textbf{Out-of-distribution performance and data efficiency.}
    \textbf{(a)} Unseen objects and distractors. \textbf{(b)} Training on $50\%$ of
    the data. Each pair of bars is labelled with the change in task completion.
    \ourexp{} stays ahead of both \maniflow{} and \pibaseline{} under
    distribution shift and reduced data. Seen ``put plate on rack'' numbers are higher than \Cref{fig:real_main} due to evaluating with 1 plate to put in the rack instead of up to 2.}
    \label{fig:real_gen}
\end{figure}

Next, we test \emph{generalization} and \emph{demonstration-efficiency} in the real world.
We re-evaluate the same models on three tasks from \Cref{sec:exp:real} but with out-of-distribution
objects, under extreme clutter visual clutter and distractors, and after retraining on only half the real-world
demonstrations.
See \Cref{fig:plate-rack-conditions} for an example and \Cref{fig:appdx:simple_tasks} for more.

\textbf{\ours{} is robust to visual distribution shift.} 
In \Cref{fig:real_gen}(a) we plot \emph{performance drop} of each method after significant visual distribution shift.
\ourexp{} gives up $4.7$ points on average at full joint generation and $4.1$ action-only, while \maniflow{}, the strongest baseline originally, drops $26.7$ despite having access to depth.
\fastwam{} holds its performance too, but only because it had almost none to give up, and \pibaseline{} loses $25.6$ points on the difficult, unseen soft bag.
Since \ourexp{} in both action-only and full joint prediction modes perform similarly well and both are significantly better than \maniflow{} with depth,
\ourexp{}'s performance improvement over other baselines is explained by the superior representation learned during WAM \emph{pre-training} with joint action-RGB-DINO-3D latent prediction.
We ablate training objectives further in \Cref{sec:additional visual inputs}.

\textbf{Superior data efficiency.} In \Cref{fig:real_gen}(b) we see that when fine-tuning on half the real-world data, \ourexp{} maintains the highest success rates, even in action-only prediction mode.
In fact, \ourexp{} with full-joint prediction trained with just half the data still outperforms all baselines trained on the full dataset, and action-only on half the data matches \pibaseline{} fine-tuned on the full set.
The world-action objective
supplies supervision that would otherwise have to come from more
demonstrations. 
We next detail simulation results, which also further test generalization and data efficiency. 
\subsection{\ref{q3}: Large-scale comparison of \ours{} against many VLAs and WAMs}
\begin{figure}[t]
\centering
    \begin{minipage}[t]{0.405\textwidth}
        \vspace{0pt}
        \centering
        \captionsetup{type=table}
        \fontsize{8}{10}\selectfont
        \setlength{\tabcolsep}{2.5pt}
        \renewcommand{\arraystretch}{1.15}
        \begin{tabular}{@{}l|ccc@{}}
        \toprule
        \textbf{Method} & \textbf{Clean} & \textbf{Rand.} & \textbf{Avg.} \\
        \midrule
        \multicolumn{4}{@{}l}{\textit{VLAs}} \\
        \midrule
        X-VLA~\citeyearpar{zheng2025xvla}                                 & 72.9 & 72.8 & 72.9 \\
        $\pi_{0.5}$~\citeyearpar{intelligence2025pi05}                    & 82.7 & 76.8 & 79.8 \\
        Qwen-RobotManip~\citeyearpar{yuan2026qwen}                        & 93.4 & 92.5 & 93.0 \\
        Qwen-RobotManip-Context                                           & 93.7 & 94.0 & 93.9 \\
        \ourcolor{\ours{} (action-only)}                            & \textbf{94.5} & \textbf{94.6} & \textbf{94.6} \\
        \midrule
        \multicolumn{4}{@{}l}{\textit{WAMs}} \\
        \midrule
        Motus~\citeyearpar{bi2025motus}                                   & 88.7 & 87.0 & 87.8 \\
        Fast-WAM~\citeyearpar{yuan2026fast}                               & 91.9 & 91.8 & 91.8 \\
        LingBot-VA~\citeyearpar{lingbot-va2026}                           & 92.9 & 91.6 & 92.2 \\
        LingBot-VA 2.0~\citeyearpar{zhang2026native}                   & 93.8 & 93.4 & 93.6 \\
        \ourcolor{\ours{} (full joint)}                             & \textbf{94.3} & \textbf{94.8} & \textbf{94.6} \\
        \bottomrule
        \end{tabular}
        \captionof{table}{
        \textbf{RoboTwin.} Success rate (\%) over 50 tasks. Training: $2{,}500$ clean $+$ $25{,}000$ randomized demos.}
        \label{tab:robotwin_main}
    \end{minipage}
\hfill
    \begin{minipage}[t]{0.57\textwidth}
        \vspace{0pt}
        \centering
        \captionsetup{type=figure}
        \includegraphics[width=\linewidth]{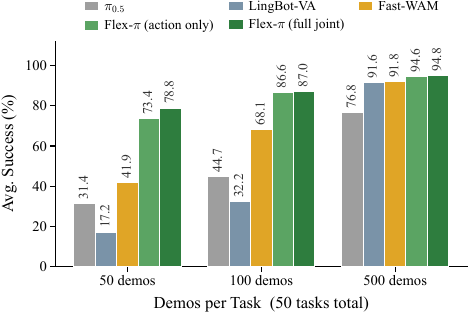}
        \caption{\textbf{RoboTwin data scaling} (domain-randomized, 50-task average). \ours{} leads at every data scale in both modes, with the largest margin at low data; baselines only close the gap at 500 demos per task.}
        \label{fig:robotwin scaling}
    \end{minipage}
\end{figure}
\textbf{RoboTwin.}
We start simulation comparisons with RoboTwin~\citep{chen2025robotwin}, on a 50-task comparison.
\Cref{tab:robotwin_main} reports success rate under the standard clean background and domain-randomized evaluation settings, we fine-tune with one checkpoint on both datasets and all tasks.
\ourexp{} leads with $94.6\%$ action-only against $93.9\%$ for the strongest VLA baseline, Qwen-RobotManip-Context~\citep{yuan2026qwen}, which uses a ${\sim}38,100$-hour pre-training corpus $76\times$ larger than ours.
Full joint prediction also outperforms every WAM baseline, including Motus~\citep{bi2025motus} and the strongest, LingBot-VA 2.0~\citep{zhang2026native}.
Interestingly, in these results, both \ourexp{} full joint and action-only perform similarly, indicating potential near-saturation of this benchmark.

We next study \textbf{data scaling} against \pibaseline{}, \lingbot{}, and \fastwam{} by varying the number of training demonstrations per task, using $N \in \{50, 100, 500\}$.
We report 50-task average success rates in \Cref{fig:robotwin scaling}, showing that \ourexp{} is significantly more data-efficient than the baselines: in lower demonstration regimes, \ourexp{} outperforms all three by $\mathbf{1.9{-}4.5\times}$. 
Importantly, full joint prediction outperforms at all data budgets, indicating that generating visual futures at inference time improves data efficiency.

\setlength{\columnsep}{16pt}
\begin{wraptable}[18]{r}{0.33\textwidth}
    \centering
    \vspace{-1em}
    \renewcommand{\arraystretch}{1.15}
    \resizebox{\linewidth}{!}{%
    \begin{tabular}{@{}l|c@{}}
        \toprule
        \textbf{Method} & \textbf{LIBERO} \\
        \midrule
        \multicolumn{2}{@{}l}{\textit{VLAs}} \\
        \midrule
        $\pi_{0.5}$~\citeyearpar{intelligence2025pi05}                    & 96.9 \\
        GR00T-N1~\citeyearpar{bjorck2025gr00t}                            & 93.9 \\
        OpenVLA-OFT~\citeyearpar{kim2025oft}                              & 97.1 \\
        MolmoAct2-Think~\citeyearpar{fang2026molmoact2}                         & 98.1 \\
        Qwen-RobotManip~\citeyearpar{yuan2026qwen}                        & 99.1 \\
        Qwen-RobotManip-Context                                           & \textbf{99.2} \\
        \ourcolor{\ours{} (action-only)}                            & 98.4 \\
        \ourcolor{\ours{}$^{*}$ (action-only)}                      & 98.7 \\
        \midrule
        \multicolumn{2}{@{}l}{\textit{WAMs}} \\
        \midrule
        LingBot-VA~\citeyearpar{lingbot-va2026}                           & 98.5 \\
        Fast-WAM~\citeyearpar{yuan2026fast}                               & 97.6 \\
        \ourcolor{\ours{} (full joint)}                             & 98.5 \\
        \ourcolor{\ours{}$^{*}$ (full joint)}                       & \textbf{99.2} \\
        \bottomrule
    \end{tabular}}
    \vspace{-0.5em}
    \caption{\textbf{LIBERO success rates.} Per-suite results are in \Cref{tab:libero_flex}. }
    \label{tab:libero_main}
\end{wraptable}
\setlength{\columnsep}{10pt}
\textbf{LIBERO.}
In LIBERO, we test the model's \emph{action-fitting capacity} (since typical evaluations contain no train/test split) against 21 other VLA and WAM baselines.
We fine-tune a single \ourexp{} checkpoint on the four standard evaluation suites (Spatial, Object, Goal, and Long; $10$ tasks each, $50$ demonstrations per task) and evaluate it with $50$ rollouts per task, reporting success rates in \Cref{tab:libero_main}; \Cref{sec:appdx:libero_setup} details how we obtain the pointmap stream in simulation and the evaluation protocol we follow. %

In LIBERO, we observe that stream dropout (\Cref{sec:method:dropout}) unsurprisingly slightly hurts action overfitting capacity. 
Therefore, we also report \ourexp{}$^{*}$ numbers, which come from fine-tuning without stream dropout.
We see full joint prediction slightly improves performance in both cases, and \ourexp{}$^{*}$ with full joint prediction matches the best baseline, Qwen-RobotManip-Context (pre-trained with $76\times$ more data) while outperforming all other VLA and WAM baselines in each respective category.

\textbf{LIBERO-Plus.}
We also evaluate on LIBERO-Plus~\citep{fei25libero-plus}, which tests generalization in LIBERO along seven axes; Appendix \Cref{tab:libero_plus} reports the per-perturbation breakdown over all $10{,}030$ perturbed tasks.\footnote{A known issue in the released code appends perturbation metadata to the instructions; we run \ourexp{}, $\pi_{0.5}$, and Fast-WAM checkpoints with it fixed, and quote the other baselines as published (\Cref{sec:appdx:libero_setup}).}
\ourexp{} reaches $88.6\%$ in \emph{both} deployment modes, ahead of $\pi_{0.5}$ at $85.7\%$ and within $0.4$ points of Qwen-RobotManip ($89.0\%$); only its context-conditioned variant scores higher ($91.4\%$).
Qwen uses $76\times$ more pre-training data than \ourexp{}---without it, Qwen reaches only $78.3\%$, $10.3\%$ below \ourexp{}.
Interestingly, \ourexp{}'s full joint generation works better on robot, language, noise, and layout generalization axes while action-only works better under camera, lighting, and background shifts. 

\subsection{\ref{q4}: Impact of \ours{}'s Additional Inputs and Outputs in Policy Performance}
\label{sec:additional visual inputs}
Finally, we perform ablation studies on additional input modalities, cross-modality forcing, output stream generation, and the reduction of flow-matching inference steps.
All three ablations share the same setup: five RoboTwin tasks (listed in \Cref{sec:appdx:addtl_exps}) with domain randomization, $50$ demonstrations per task, and $5$ epochs from scratch.

\begin{figure}[t]
    \centering

    \begin{subfigure}[t]{0.46\linewidth}
        \centering
        \includegraphics[width=\linewidth]{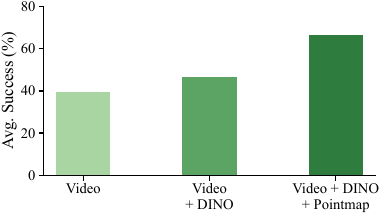}
        \caption{Input streams.}
        \label{fig:training_modalities}
    \end{subfigure}
    \hfill
    \begin{subfigure}[t]{0.50\linewidth}
        \centering
        \includegraphics[width=\linewidth]{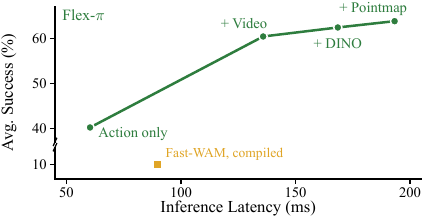}
        \caption{Latency vs.\ success.}
        \label{fig:latency_vs_success_robotwin}
    \end{subfigure}

    \caption{
    \textbf{RoboTwin ablations and system tradeoffs} (\Cref{sec:appdx:addtl_exps}).
    \textbf{(a):} which visual streams the model observes, added cumulatively. For each ablation, all available streams are predicted at inference.
    \textbf{(b):} with RGB-only input, generating more streams trades latency for success, from one checkpoint throughout.
    }
    \label{fig:robotwin_ablations_latency}
\end{figure}

\textbf{Input Modalities.} First, we ablate each \emph{input} modality \ourexp{} conditions on, training with video only, video + DINO, and all three visual streams (video + DINO + pointmap).
In \Cref{fig:training_modalities}, adding DINO to video increases success by $6.8\%$, and adding pointmaps on top of both increases it by a further $20\%$.
DINO adds object-centric structure and pointmaps add explicit 3D geometry, significantly improving policy performance over video-only predictions during the training process.

\textbf{Output Modalities.} Next, given \emph{the same pre-trained checkpoint} and RGB-only input, we vary which streams \ours{} \emph{generates} at inference (\Cref{fig:latency_vs_success_robotwin}).
The action-only path reaches $40.2\%$ at ${\sim}60$\,ms, undercutting even a PyTorch-compiled Fast-WAM ($90$\,ms) at four times its success rate ($10.0\%$). Also generating video lifts success to $60.4\%$, and adding DINO and pointmaps reaches $63.8\%$ at ${\sim}193$\,ms.
A single checkpoint thus spans more than a $3\times$ latency range and $24$ points of success, leaving the operating point to be chosen at deployment.

\textbf{Cross-Modality Forcing.} \Cref{fig:forcing_ablation} ablates cross-modality forcing, i.e., training to generate visual modalities not present in the input mask (\Cref{sec:method:dropout}).
Removing cross-modality forcing actually \emph{hurts} success rates by $21\%$:
\begin{wrapfigure}{r}{0.3\textwidth}
    \vspace{-1.2em}
    \includegraphics[width=\linewidth]{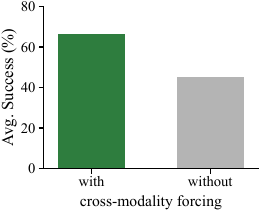}
    \caption{\textbf{Cross-modality forcing} on RoboTwin. Both models observe all three streams; only the training rule differs.}
    \label{fig:forcing_ablation}
\end{wrapfigure}
The benefit of cross-modality forcing is therefore not only robustness to missing sensors at inference time, but also that requiring each modality to be predictable from the others encourages \ourexp{} to construct a representation in which appearance, geometry, and semantics are mutually predictive.
This stronger representation results in better action prediction.
Appendix \Cref{fig:appdx:agibot_streams} shows what this looks like at inference: with the pointmap withheld from the input, the same checkpoint still generates scene geometry matching what it produces when all three streams are observed.

\textbf{Few-Step Inference.} We also ablate the number of Euler steps $K$ during flow matching.
Sweeping it on the same checkpoint, action-only success peaks at $K{=}4$ and stays within $1.0$ point of that peak for every $K\ge2$, so we used $K{=}4$ throughout our experiments, at ${\sim}60$\,ms action-only and ${\sim}193$\,ms full joint  inference time. See \Cref{sec:appdx:steps} for results and more details.

\section{Conclusion and Limitation}
\label{sec:conclusion}

We introduce \ours{}, a world-action model that embeds RGB, 3D, and DINO
semantics into a shared latent space, yielding a single checkpoint that supports
any combination of input and output streams at inference without the need for additional sensor modalities or training new visual priors.
Our results demonstrate that it outperforms the strongest baselines on real robot hardware, while remaining faster than all of them in action-only mode. 
Ablations demonstrate that its demonstration efficiency and generalization capabilities come from the WAM training objective on all visual streams.

\textbf{Limitations.} Although well-optimized, the additional modalities and cross-modality forcing mean that \ours{} takes longer to converge, requiring at least 10 epochs of fine-tuning on our real-world tasks. 
Furthermore, the joint generation mode of all output modalities is still slower than parameter-comparable VLAs.
LIBERO-Plus experiments, where the only baselines still ahead of \ours{} pair a strong VLM backbone with $76\times$ more robot pre-training data, demonstrate that having strong semantic reasoning capabilities and access to significantly more robot data would further improve \ours{}, albeit while also increasing needed computation for pre-training.
We leave addressing these challenges to future work and encourage the community to build upon \ours{}.

\section*{Acknowledgments}
We thank Katherine Liu for providing valuable writing feedback.
Additionally, we thank the University of Washington Hyak and Tillicum high-performance computing clusters for providing us with compute resources.
Ge Yan, Jesse Zhang, and Dieter Fox acknowledge compute and funding support by the Toyota Research Institute, and support by the Cross-Pacific AI Initiative from Amazon.

\bibliography{refs}  %
\bibliographystyle{iclr2026_conference}

\clearpage
\appendix

\raggedbottom

\etocdepthtag.toc{appendix}
\etocsettagdepth{mainmatter}{none}
\etocsettagdepth{appendix}{subsection}
\renewcommand{\contentsname}{Appendix Contents}
\tableofcontents

\clearpage

\section{Model Architecture Details}
\label{sec:appdx:architecture}
 
\subsection{Proprioception Encoding}
\label{sec:appdx:proprioception}
 
\textbf{Canonical state layout.}
All embodiments share a single $32$-dimensional proprioception vector. Each arm owns $16$ of the $32$ channels: the end-effector position ($3$), its orientation in a continuous 6D rotation representation ($6$), the gripper state ($1$), and the arm's joint positions ($6$). The vector is grouped by \emph{field} rather than by arm, so the two end-effector poses come first, then the two grippers, then the two joint blocks.
\begin{center}
\begin{tabular}{ll@{\qquad}ll}
\toprule
Slots & Contents & Slots & Contents \\
\midrule
$0$--$2$, $9$--$11$  & L/R end-effector position   & $18$, $19$   & L/R gripper \\
$3$--$8$, $12$--$17$ & L/R rotation (6D)           & $20$--$31$   & L/R joint positions \\
\bottomrule
\end{tabular}
\end{center}
Rather than left-packing each robot's native state, we scatter its channels into the fixed slots above, zero-filling unused slots and storing a per-dimension padding mask. This keeps shared channels at consistent indices and allows the same pre-trained proprioception encoder to transfer across embodiments without reshaping. LIBERO is the exception: it uses a $3$-D axis-angle rotation in slots $3$--$5$, whereas other embodiments use the first three components of a 6D rotation.

At fine-tuning time, RoboTwin's native $14$-D bimanual state is mapped into the shared $32$-D layout, allowing the AGIBOT World encoder to be reused directly. LIBERO occupies $8$ slots, while the real-robot YAM platform (\Cref{sec:appdx:yam}) uses all $32$.

Normalization is embodiment-specific. Pre-training and real-robot runs linearly map each non-rotation channel to $[-1,1]$ using per-episode $1$st/$99$th-percentile bounds, aggregated across episodes by min/max. The 6D rotation slots are left unchanged and orthonormalized with Gram--Schmidt at decode time. LIBERO instead z-scores all channels, including its axis-angle rotation, since it does not use a 6D rotation representation.

\textbf{Encoder and conditioning.}
Only the state at the current timestep, $s_t$, is encoded; the model receives no proprioception history.
The encoder is a single linear layer $\mathbb{R}^{32} \rightarrow \mathbb{R}^{4096}$ mapping $s_t$ to the width of the \texttt{umT5} text tokens, and thus emitting exactly one token.
That token is then \emph{appended to the language token sequence itself}, with the text attention mask extended by one entry.
Proprioception consequently reaches $\latentmodel$ through the same cross-attention pathway as the instruction $l$, as one additional conditioning token rather than a separate input branch or an AdaLN offset, so every stream, and the action expert, attend to the robot's state exactly as they attend to language.
 
\subsection{Action Representation and Chunk Horizon}
\label{sec:appdx:actions}
 
\textbf{Representation.}
Actions reuse the $32$-dimensional canonical layout of \Cref{sec:appdx:proprioception}, so a given slot denotes the same physical quantity whether it is being observed as state or predicted as an action.
The two differ in frame of reference: proprioception is always absolute, whereas on the real robot every step of an action chunk is expressed \emph{relative} to a single anchor: the state observed at the chunk's first timestep.
Under that anchor, end-effector targets become body-frame relative poses $R_{\text{base}}^{\top}(p - p_{\text{base}})$ with rotations reparameterized as $\mathrm{rot6D}(R_{\text{base}}^{\top} R)$, joint angles become scalar displacements, and the two gripper channels stay absolute (continuous in $[0,1]$ on the real robot; binary in LIBERO).
Anchoring all $H$ steps to one pose, rather than each step to its predecessor, keeps targets from accumulating integration error along the chunk and makes the representation invariant to where in the workspace the motion begins.
LIBERO keeps its native per-step operational-space deltas inside the same slots, so the canonical layout fixes \emph{which} channel each slot carries, while the choice of reference frame remains a per-embodiment property of the data.
 
\textbf{Chunk horizon.}
Each training sample spans $33$ consecutive timesteps, yielding an action chunk of $H = 32$ actions.
The RGB and pointmap streams are subsampled from that same window at a stride of $4$ into $9$ frames, which the VAE's $4\times$ temporal compression turns into $3$ latent frames: the current observation and two futures. The DINO stream is taken at those same three timestamps.
The action chunk and the generated visual future therefore cover an identical horizon: the futures a given action attends to under $\outputmask$ are always the futures contemporaneous with it.
How much of that chunk is executed before re-planning is a deployment choice rather than a property of the representation: on the real robot we run all $32$ steps open-loop (\Cref{sec:appdx:yam}), whereas in LIBERO we re-plan every $10$ steps. 
 
\textbf{Tokenization and head.}
We do not discretize actions. Each of the $H$ timesteps is embedded by a single linear layer into one token at the action expert's width (\Cref{sec:appdx:adapters}), giving $32$ action tokens that are denoised jointly under the flow-matching objective of \Cref{eq:total-loss}; a single linear layer maps each denoised token back to the $32$-dimensional action space.
Both are randomly initialized at the start of pre-training, being the two modules with no counterpart in a video model; because the canonical layout keeps their shapes fixed across embodiments, both are then carried over intact when a pre-trained checkpoint is fine-tuned.
 
\subsection{Action Expert Initialization and Per-Modality Adapters}
\label{sec:appdx:adapters}
 
\textbf{Action expert initialization.}
The action expert matches the video expert in head count ($24$), per-head dimension ($128$), and depth ($30$ blocks); these three must agree so that the two streams' queries, keys, and values stay shape-compatible when concatenated for MoT joint attention.
Its residual and feedforward widths are smaller, $d_a = 1024$ and $4096$ against $d_v = 3072$ and $14336$, which makes a direct weight copy impossible.
We therefore initialize by \emph{resampling}: each backbone tensor is resized one axis at a time to its target shape by 1D linear interpolation, and any tensor whose fan-in was reduced is rescaled by $\sqrt{d_v/d_a}$ so that activation variance is preserved at initialization.
Only the action encoder and the action output head, which have no counterpart in a video model, are randomly initialized.
 
\textbf{Per-modality adapters.}
The shared visual trunk operates at the single width $d_v$, so every stream needs a mapping into and out of it.
Pointmaps are encoded by the same frozen VAE as RGB and therefore already lie in the video latent space; their adapter is a private copy of the video expert's patch embedding and unpatchify head, initialized from the pre-trained \texttt{Wan-2.2} weights so the pointmap stream inherits the video prior before specializing.
DINO features do not share that space, so the DINO stream instead applies a LayerNorm to the raw features followed by a single linear layer in and a single linear layer out, both Xavier-initialized, between the folded DINO width and $d_v$ (below).
The action stream likewise uses one linear layer in and one out, at its own width $d_a$.
 
\textbf{DINO feature folding.}
A DINOv3 encoder emits one $768$-dimensional token per image patch, and because \ours{} \emph{generates} future DINO features as well as consuming current ones, DINO tokens can become quite compute-intensive in the observation and in generation.
We therefore apply a pixel-unshuffle ~\citep{shi2016pixleshuffle} (space-to-channel) fold of factor $f = 2$ to each view's native patch grid before the adapter: each $2 \times 2$ block of neighboring patches is concatenated along the channel axis, so the encoder's $14 \times 14$ grid of $768$-dimensional tokens becomes a $7 \times 7$ grid of $f^2 \cdot 768 = 3072$-dimensional tokens per view.
The rearrangement is exactly invertible, so no feature content is lost; the model simply predicts each $2 \times 2$ neighborhood jointly within one token instead of across four.
Token count per view drops $f^2 = 4\times$, i.e.\ $75\%$ fewer DINO tokens in both the attention sequence and the generation target, and the RoPE grid is the post-fold grid so positions remain one-per-token.
 
\subsection{$x$-Prediction for the Folded DINO Stream}
\label{sec:appdx:dinoparam}
 
\textbf{Why velocity prediction is hard.}
Folding raises each DINO token to $f^2 \cdot 768 = 3072$ dimensions, the same as the feedforward hidden dim $d_v$.
\citet{li2025back} demonstrate that $x$-prediction~\citep{salimans2022progressive} works better than standard velocity prediction (\Cref{eq:fm-loss}) when the data dimension is large, and we also found this to empirically be the case for our setting.
Only the folded DINO stream sits at this ratio: the video and pointmap latents are at $192/3072 \approx 0.06$ and the action stream between $0.01$ and $0.03$ of its own width $d_a$, all far from the regime where $x$-prediction is required, so they use standard flow matching velocity prediction.
 
\textbf{Clean-feature prediction.}
The DINO head therefore estimates the clean features $\hat{d}_{t+1}$, and we convert that estimate to a velocity for the standard flow matching loss.
The linear path is $d^{\tau}_{t+1} = \tau d_{t+1} + (1-\tau)\epsilon$, where $\tau$ is the flow timestep $\in [0, 1]$, and the identity $d_{t+1} - d^{\tau}_{t+1} = (1-\tau)(d_{t+1} - \epsilon)$ gives us the velocity $\hat{v}$ by:
\begin{equation}
    \hat{v} \;=\; \frac{\hat{d}_{t+1} - d^{\tau}_{t+1}}{1 - \tau}.
    \label{eq:dino-x0}
\end{equation}
\Cref{eq:fm-loss,eq:total-loss} are therefore untouched: \Cref{eq:dino-x0} converts the $x$-prediction into a velocity, which is used for the loss and for Euler integration exactly as a velocity head's output would; only how the head's output is read changes.
See Table 1 of \citet{li2025back} for a comparison of noise, $x$, and velocity prediction.
 
\subsection{Stream Masking: Attention Rules}
\label{sec:appdx:regimes}
 
\textbf{Sampling.}
$\inputmask$ and $\outputmask$ are drawn independently per sample, each entry Bernoulli$(0.5)$, with rejection sampling on $\inputmask$ so that at least one visual stream is always observed.
A stream absent from $\inputmask$ has its current-timestep tokens zeroed rather than removed: the sequence layout is fixed, so tensor shapes stay identical across a batch.
Cross-modal prediction is enabled for all three visual streams, so a presence-dropped stream is still denoised rather than dropped from its loss.
 
\textbf{Attention rules.}
Write $\text{obs}$ for the current-timestep tokens of the observed streams, $X_{t+1}$ for a future visual stream, and $a$ for the action tokens. Within the joint-attention layers:
\begin{itemize}[leftmargin=1.4em,itemsep=1pt,topsep=2pt]
    \item $X_{t+1} \rightarrow \text{obs}$ and $a \rightarrow \text{obs}$ always, for \emph{every} observed stream, not only a stream's own modality;
    \item $X_{t+1} \leftrightarrow Y_{t+1}$ iff $m^\text{out}_X = m^\text{out}_Y$, so $\outputmask$ splits the futures into two groups and removes attention only \emph{between} them. Two streams the action does not read therefore remain visible to one another;
    \item $a \rightarrow X_{t+1}$ iff $m^\text{out}_X = 1$, and one-way only. No token ever attends to the action stream, so it can be dropped entirely at inference.
\end{itemize}
The action stream is never masked, and $\outputmask$ enters no term of \Cref{eq:total-loss}: every head is denoised and supervised on every draw, keeping cross-modality forcing well-posed.
 
\textbf{Scope.}
These rules govern the middle $16$ of the $30$ MoT blocks; the outer blocks attend within each stream independently (\Cref{sec:method:streams}).
Language and proprioception reach every token by cross-attention regardless of either mask, so even a fully presence-dropped stream still receives the task conditioning.
 
\textbf{Training versus inference.}
At training, $\outputmask$ is purely an attention pattern and all four streams are generated on every sample.
At inference it additionally selects which futures are \emph{computed}: a stream the action does not read is dropped from the sequence outright, producing the action-only speedup.

\section{Pre-training Data}
\label{sec:appdx:pretraining}

\textbf{Episode sampling and re-segmentation.}
From $100$ tasks of AGIBOT World-Beta~\citep{contributors2024agibotworldrepo} we draw the first $285$ episodes of each task, taking all of them where a task holds fewer, which is the case for $12$ tasks; this yields roughly $500$ hours in total.
AGIBOT World stores one episode per \emph{full} task execution, whereas its language annotations label the individual action segments \emph{within} an episode.
We therefore re-segment every sampled episode at those annotation boundaries, so that each training sample is a sub-episode paired with the instruction that actually describes the motion it contains, rather than with a whole-task label.

\textbf{Large-scale depth labeling.}
AGIBOT World does release sensed depth, but only for the head view, so the two wrist views would have no geometry at all.
The head stream is also uneven for our purpose, dropping large regions to holes and returning noisy values elsewhere, and that error would propagate straight into the pointmaps the VAE has to encode.
We therefore leave it aside and annotate all three views of the sampled episodes offline from RGB alone with a monocular metric-depth estimator (Depth Anything~3)~\citep{lin2025depth}, lift the result to pointmaps with only the camera intrinsics (no calibrated extrinsics needed), and tile them into the same three-view composite canvas we use on the real robot (\Cref{sec:appdx:yam}).

\begin{figure}[p]
    \centering
    \includegraphics[width=\linewidth]{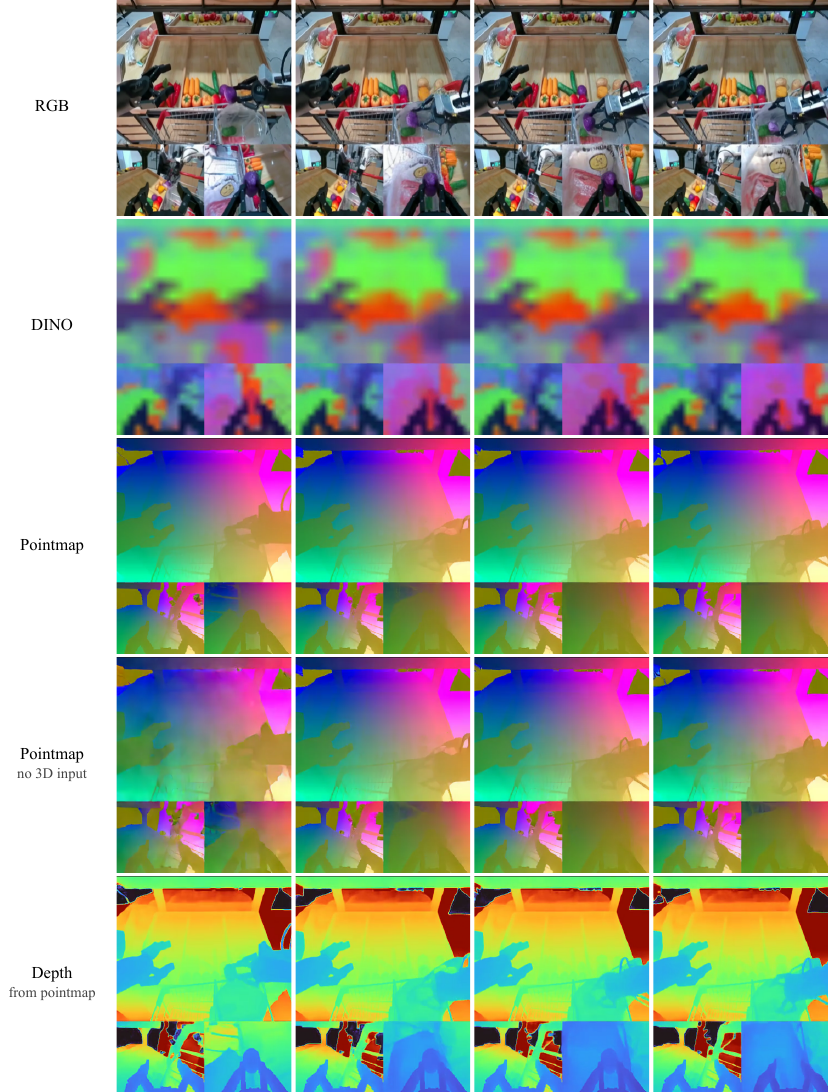}
    \caption{\textbf{Generated streams} on AGIBOT World. Examples of the visual streams a single \ours{} checkpoint generates, at sampled frames (left to right); each panel is the three-view composite canvas, head camera above and the two wrist cameras below. The first three rows generate with every stream observed. The fourth repeats the episode with the pointmap dropped from the input, so the geometry has to come from RGB and DINO alone (cross-modality forcing in \Cref{sec:method:dropout}), yet the scene structure still looks good. The last row renders the third row's generated pointmap as depth, which is easier to see; it is not a separate output.}
    \label{fig:appdx:agibot_streams}
\end{figure}

\section{Real-World Robot Platform (YAM)}
\label{sec:appdx:yam}
 
\textbf{Hardware.}
We run all real-robot experiments on a stationary bimanual YAM setup: two $6$-DoF arms, each with a single-DoF parallel gripper, for $14$ commanded degrees of freedom in total, filling the $32$-dimensional canonical layout of \Cref{sec:appdx:proprioception} exactly.
We mount three calibrated stereo cameras, a ZED~2i overhead and a ZED~Mini on each wrist, each returning an RGB image and a \emph{sensed} metric depth map; we lift depth to pointmaps using the per-camera intrinsics shipped with every frame.
 
\textbf{Observations and actions.}
We tile the three views into one composite canvas, overhead on top and the two wrists side by side beneath it, so all three views cost a single VAE pass. The layout is checkpoint-specific and read from that checkpoint's configuration at deployment.
We record demonstrations and command the arms at $30$\,Hz, querying the policy once per chunk. Each chunk is conditioned on a single observation frame, with no visual or proprioceptive history, and the policy predicts $H{=}32$ actions alongside a $9$-frame window per generated visual stream: the first frame is the current observation and the remaining $8$ are future.
Proprioception is the absolute $32$-D state read from the arms; actions use the body-frame relative representation of \Cref{sec:appdx:actions}, converted back to absolute targets against the current state at execution time. States and actions are normalized the same way (\Cref{sec:appdx:proprioception}).
 
\textbf{Deployment.}
We serve the policy over a local socket on a single RTX~5090. The robot executes all $32$ predicted steps before the next chunk is planned, so the policy re-plans every $1.07$\,s. Our client is synchronous, so the arms hold position at each chunk boundary while the next chunk is computed; inference cost therefore surfaces as idle time between chunks rather than a slower control loop, and reported episode times include it.
The input and output stream masks are runtime arguments rather than architectural choices, so the action-only and full-joint conditions come from one checkpoint invoked with different flags, not separately trained models.

\section{Real World Experiment Criteria}
\label{app:data-real-world-eval-criteria}
Every rollout is scored by a human evaluator against the partial-credit rubric of
\Cref{tab:appdx:rubrics}. The reported \emph{score} is the credit earned normalized by
the maximum attainable, not a binary success rate.

\begin{table}[t]
\caption{\textbf{Real-world scoring rubrics.} Stages are listed in execution order.
\emph{Self-Repair Gripper} and \emph{Soft-Bag Zipping} are strictly sequential---each
stage presupposes the ones above it---so for those two we also report the fraction of
rollouts that clear every stage (\Cref{fig:selfrepair_results}).}
\label{tab:appdx:rubrics}
\vspace{4pt}
\centering
\small
\setlength{\tabcolsep}{4pt}
\renewcommand{\arraystretch}{1.1}
\begin{tabular}{@{}lr@{}}
\toprule
\textbf{Stage} & \textbf{Points} \\
\midrule
\multicolumn{2}{@{}l}{\textit{Put Plate on Rack} --- $N$ plates start on the table beside the rack; scored per plate} \\
\quad Push one edge of the plate down and grasp it from the tilted edge & $0.5$ \\
\quad Place it in the rack in any condition, including an occupied or misaligned slot & $0.25$ \\
\quad Place it cleanly into an empty slot (in addition to the credit above) & $0.25$ \\
\quad \textit{Maximum, per plate} & \textit{1.0} \\
\midrule
\multicolumn{2}{@{}l}{\textit{Sort Utensils} --- two utensils, several plates of different colors, and a container} \\
\quad Place one utensil onto any plate & $0.5$ \\
\quad That plate is the one of the specified color & $0.5$ \\
\quad Insert the second utensil vertically into the container & $1.0$ \\
\quad \textit{Maximum} & \textit{2.0} \\
\midrule
\multicolumn{2}{@{}l}{\textit{Kitchen Organization} --- bowls and a plate on the table, a rack within reach of both arms} \\
\quad Place the bowls onto the rack & $1.0$ \\
\quad Hand the plate from one arm to the other without dropping it & $0.5$ \\
\quad Insert the plate correctly into a rack slot & $0.5$ \\
\quad \textit{Maximum} & \textit{2.0} \\
\midrule
\multicolumn{2}{@{}l}{\textit{Self-Repair Gripper} --- gripper, screw holder, and powered driver at randomized} \\
\multicolumn{2}{@{}l}{positions on the right; vegetable and bucket on the left} \\
\quad Reach the pose from which the repair can begin & $0.25$ \\
\quad Pick up the replacement gripper & $0.25$ \\
\quad Insert the gripper into its holder & $0.5$ \\
\quad Pick up the screw & $0.5$ \\
\quad Insert the screw into the mounting hole & $0.25$ \\
\quad Pick up the driver & $0.25$ \\
\quad Drive the screw home & $0.75$ \\
\quad Place the vegetable into the bucket & $0.25$ \\
\quad \textit{Maximum} & \textit{3.0} \\
\midrule
\multicolumn{2}{@{}l}{\textit{Soft-Bag Zipping} --- a closed fabric pouch and one or two pens, all at randomized positions} \\
\quad Close the gripper on the zipper pull, to open & $1.0$ \\
\quad Draw the slider far enough that the mouth admits a pen & $1.0$ \\
\quad Hold the mouth open wide enough to insert a pen without it touching the rim & $0.25$ \\
\quad Place every pen inside, divided evenly among the pens present & $0.25$ \\
\quad Close the gripper on the zipper pull again, to close & $1.0$ \\
\quad Draw the slider to its closed end stop & $1.0$ \\
\quad \textit{Maximum} & \textit{4.5} \\
\bottomrule
\end{tabular}
\end{table}

\section{LIBERO Setup and Evaluation Protocol}
\label{sec:appdx:libero_setup}

\textbf{Camera layout.}
LIBERO provides an agentview and a single wrist camera, one fewer than the three-slot
composite of \Cref{sec:appdx:yam}. We fill the right-wrist slot with zeros in RGB and
depth alike rather than special-casing the model, so one encoder, one composite geometry,
and one set of layout constants serve LIBERO, RoboTwin, and the real robot.

\textbf{Depth and pointmaps.}
During fine-tuning, LIBERO's depth is rendered from the simulator rather than estimated. The released demonstrations carry RGB
only, so instead of annotating them the way we annotate AGIBOT World
(\Cref{sec:appdx:pretraining}), we replay each one in the simulator with the depth pass
enabled and re-render it at $512{\times}512$, capturing RGB and depth together for both
cameras. MuJoCo returns a normalized $z$-buffer, which we convert to metric depth with
robosuite's~\citep{robosuite2020} camera utilities and store as \texttt{uint16} millimetres, losslessly
encoded alongside the video. Intrinsics are read once from the scene rather than per
task: both cameras are declared in robosuite's XML, so a single $K$ per camera holds
across every task and suite, and we rescale it to the resolution each stream is consumed
at. Pointmaps are then unprojected with $K$ alone, in the camera frame and needing no
extrinsics, and clipped at $2$\,m, well outside the working volume of a LIBERO tabletop.
The same code runs at evaluation, so training targets and test-time observations come
from one renderer.

\textbf{States and actions.}
LIBERO's $8$-dimensional state (end-effector position, a $3$-D axis-angle orientation,
and two gripper channels) and its $7$-dimensional action (per-step operational-space
deltas plus the gripper) are scattered into the canonical $32$-dimensional layout of
\Cref{sec:appdx:proprioception}, with the axis-angle in slots $3$--$5$. The platform is
single-armed and exposes no joint readings, so it occupies $8$ slots and zero-pads the
other $24$. Every channel is $z$-scored.

\textbf{Evaluation.}
We fine-tune across all four suites at once rather than one model per suite, and evaluate
each of the $40$ tasks over $50$ rollouts, so every average in \Cref{tab:libero_flex}
rests on $2{,}000$ episodes. We adopt the step budgets the LIBERO leaderboard is scored
under---$220$/$280$/$300$/$520$ for Spatial/Object/Goal/Long---and use each task's
language annotation verbatim as the instruction. A call predicts $H{=}32$ actions with
$K{=}4$ Euler steps, of which $10$ are executed before re-planning.

\textbf{LIBERO-Plus.}
We run all $10{,}030$ perturbed tasks at one trial each, with a fresh environment per
task, the official initial states and success predicates, and the official
task-count-weighted \emph{Total} rather than a mean over the seven categories. Our one
addition to the simulator is the depth render pass, which changes neither the rendered
RGB nor the physics.

LIBERO-Plus perturbs seven axes, and only one of them (\emph{Language}) is meant to
alter the task instruction; the other six perturb the camera, the robot's appearance,
lighting, background, sensor noise, or object layout and leave the prompt as it is in
standard LIBERO. As reported in the benchmark's public issue tracker, the released
evaluation code does not preserve that separation: for the six non-language categories
it reconstructs the instruction from the variant filename, appending perturbation
identifiers to the prompt the policy receives. We fix the extraction so that each task
is run with its own language annotation, and leave initial states, success predicates,
task counts, and the \emph{Total} weighting as released.

On the task-count-weighted \emph{Total}, this moves \ours{} from $78.3\%$ to $88.6\%$ in
action-only mode and from $80.9\%$ to $88.6\%$ at full joint generation, $\pi_{0.5}$ from
$84.7\%$ to $85.7\%$, and Fast-WAM from $49.0\%$ to $70.8\%$, all from the same
checkpoints on the same tasks. \Cref{tab:libero_plus} reports these corrected numbers and reproduces
the remaining baselines as published: we did not re-run those, and the published work
does not report which instruction handling produced its leaderboard, so we quote them
unchanged.

\section{Baseline Implementation Details}
\label{sec:appdx:baselines}
This section records where every baseline number in the paper comes from and, for the
baselines we trained ourselves, how they were configured and deployed.

\subsection{Simulation Baseline Sources}
\label{sec:appdx:baseline_provenance}
We train no baseline on LIBERO or LIBERO-Plus. Every LIBERO baseline number
(\Cref{tab:libero_flex}) is a published figure from the original work under the same
evaluation protocol.

LIBERO-Plus needs one further distinction, because we report it under the corrected
instruction protocol of \Cref{sec:appdx:libero_setup} rather than the released one.
Neither Fast-WAM~\citep{yuan2026fast} nor $\pi_{0.5}$~\citep{intelligence2025pi05}
publishes LIBERO-Plus results at all, so the rows marked $^{\dagger}$ in
\Cref{tab:libero_plus} are our own evaluations of their released checkpoints, run
through the same corrected pipeline as \ours{} with no weights updated --- these are the
like-for-like comparisons. The remaining baselines are published figures from
\citet{fei25libero-plus}, reproduced as reported: re-running each of them over all
$10{,}030$ perturbed tasks was outside our compute budget.

The RoboTwin data-scaling experiment (\Cref{fig:robotwin scaling}) is the one simulation
setting we train ourselves, since it probes demonstration budgets no published work reports:
$\pi_{0.5}$, Fast-WAM, and LingBot-VA are each retrained at $50$ and $100$ demonstrations per
task. The $500$-demonstration point needs no retraining --- it is the setting the original
work already reports, so we quote those published figures, given per task in
\Cref{tab:appdx:robotwin_pertask}.

\subsection{Real-Robot Baseline Deployment}
\label{sec:appdx:baseline_deploy}
No published numbers exist for the YAM platform (\Cref{sec:appdx:yam}) or the five tasks of
\Cref{sec:appdx:addtl_exps}, so every real-robot baseline --- $\pi_{0.5}$~\citep{intelligence2025pi05},
ManiFlow~\citep{yan2025maniflow}, and Fast-WAM~\citep{yuan2026fast} --- is one we trained.
For a given task, every model, \ours{} included, sees an identical dataset: the same
demonstrations, DAgger corrections where the task has them, train split, and preprocessing.
Beyond the data, each baseline runs in its own native configuration: several carry
pre-trained weights trained under a particular observation layout, action parameterization,
and chunk length, which we keep at the released defaults rather than override. Real-robot
performance differences are therefore attributable to the model and training recipe, not to
the data shown.

\textbf{Task coverage.} Not every baseline runs on every task: $\pi_{0.5}$ and ManiFlow cover
all five, while Fast-WAM covers \emph{Put Plate on Rack}, \emph{Sort Utensils}, and
\emph{Kitchen Organization} only, since its performance on those three indicated it would not
reach a scoreable level on the two long-horizon dexterous tasks. \Cref{tab:appdx:task_coverage}
records the pairing; every comparison we report is against baselines actually run on that
task.

\begin{table}[h]
\caption{\textbf{Real-robot task coverage.} Which baselines were trained and evaluated on
each task.}
\label{tab:appdx:task_coverage}
\vspace{4pt}
\centering
\small
\setlength{\tabcolsep}{8pt}
\renewcommand{\arraystretch}{1.15}
\begin{tabular}{lccc}
\toprule
Task & $\pi_{0.5}$ & ManiFlow & Fast-WAM \\
\midrule
Put Plate on Rack     & \checkmark & \checkmark & \checkmark \\
Sort Utensils         & \checkmark & \checkmark & \checkmark \\
Kitchen Organization  & \checkmark & \checkmark & \checkmark \\
Soft-Bag Zipping      & \checkmark & \checkmark & \\
Self-Repair Gripper   & \checkmark & \checkmark & \\
\bottomrule
\end{tabular}
\end{table}
\textbf{Camera inputs.} The baselines do not share one view of the scene.
Fast-WAM~\citep{yuan2026fast} takes the same three-view composite canvas \ours{} does
(\Cref{sec:appdx:yam}) at $384\!\times\!320$, RGB only. $\pi_{0.5}$~\citep{intelligence2025pi05}
takes the three views as separate $224\!\times\!224$ tensors, matching its released
checkpoint. ManiFlow~\citep{yan2025maniflow} takes the three RGB views at the same resolution
plus the three depth maps, back-projected to pointmaps on device by its own encoder.
This departs from its released recipe, which conditions on a sampled point cloud. We ran
ManiFlow both ways and found that RGB plus pointmaps performs much better on our tasks, so
we use it as the stronger baseline.

\textbf{Chunk length and control rate.} Every method is queried once per chunk and the arms are commanded
at $30$\,Hz (\Cref{sec:appdx:yam}). Chunk length differs following best practices for each baseline: $\pi_{0.5}$ and ManiFlow predict
$50$ steps, Fast-WAM and \ours{} predict $32$. All methods share the same synchronous client,
so the arms hold position at each chunk boundary and inference cost surfaces as idle time
between chunks---included in wall-clock episode time for every method, though a longer
chunk pays it fewer times per episode.

\subsection{Latency Measurement Protocol}
\label{sec:appdx:baseline_latency}

\textbf{Common conditions.} All latencies are measured on the same machine and RTX 5090, at
batch size $1$, with warmup calls discarded. The unit is one policy call producing one action
chunk, not one control step, since every method is queried once per chunk
(\Cref{sec:appdx:baseline_deploy}); text encoding sits outside the timed region wherever a
method caches it per task, matching deployment.

\textbf{Each method is measured in its deployed configuration.} We do not hold optimization
level fixed across methods: the comparison reports what each policy costs as we actually run
it, not what it would cost under equal engineering effort. \ours{}'s full-joint path used 
the optimization stack of \Cref{sec:appdx:inference} to reach a deployable range. 
We did optimize Fast-WAM via adding torch compilation to make it much faster. We report these faster numbers. 

\section{Additional Experiments}
\label{sec:appdx:addtl_exps}

\subsection{\ref{q1} Real-World Evaluation: Protocol and Generalization Analysis}
\label{sec:q3}
\label{sec:real_generalization}

\begin{figure}[t]
\centering
\includegraphics[width=\linewidth]{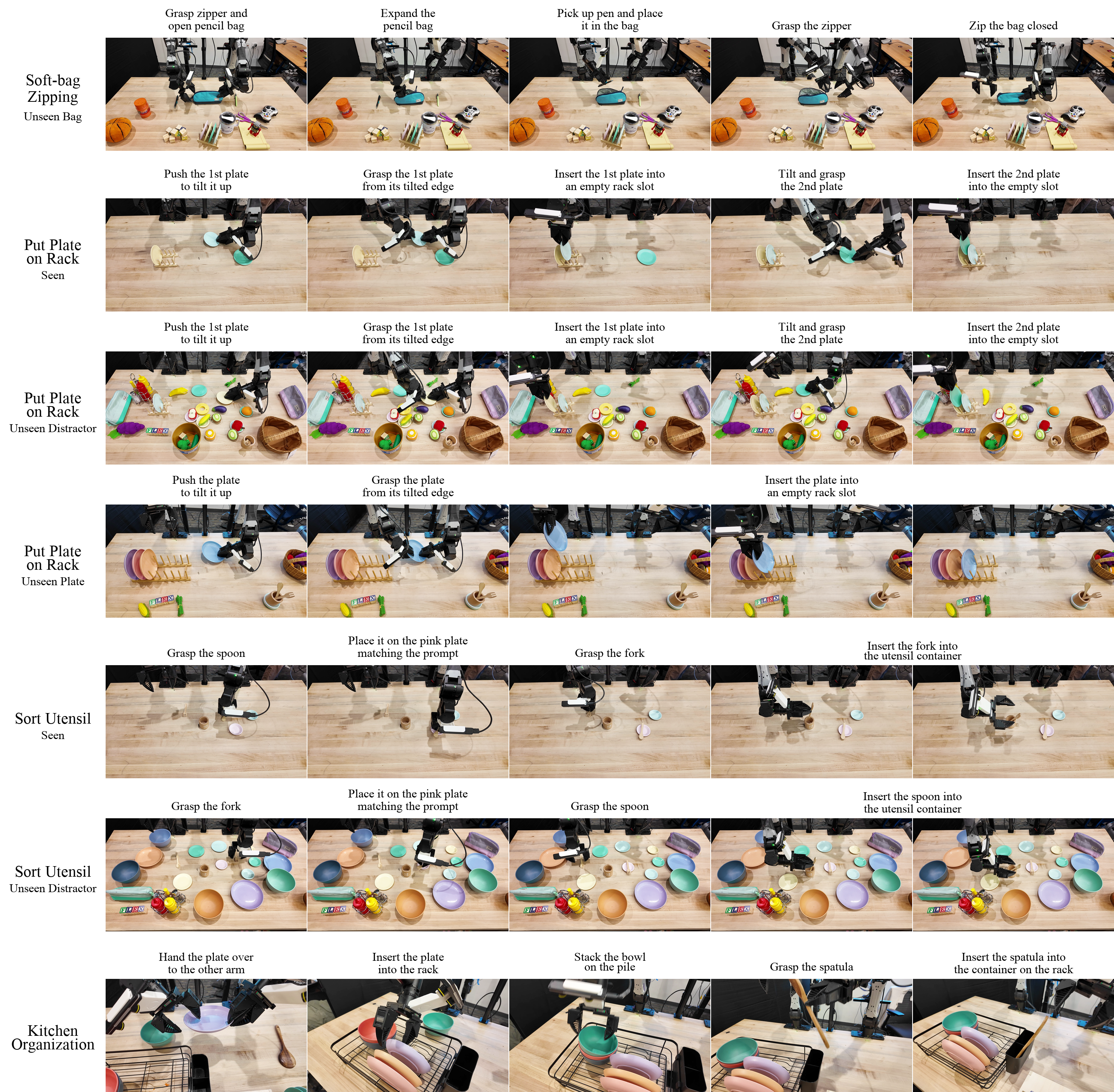}
\caption{\textbf{The remaining evaluation tasks and generalization cases.}
\emph{Put Plate on Rack} and \emph{Sort Utensils} test bimanual coordination
under sustained contact; \emph{Kitchen Organization} chains four such skills into
one long-horizon episode and is evaluated in a single setting. Rows are grouped
by task, with the seen condition first and the unseen conditions after: added
distractor objects for \emph{Put Plate on Rack} and \emph{Sort Utensils}, and an
unseen plate for \emph{Put Plate on Rack}. The top row is the unseen-bag case of
\emph{Soft-Bag Zipping}, whose seen condition appears in \Cref{fig:hard_tasks}
alongside \emph{Self-Repair Gripper}.}
\label{fig:appdx:simple_tasks}
\end{figure}

\begin{table}[t]
\caption{\textbf{Real-robot training data per task.} Hours are wall-clock
demonstration time, and every method trains on the identical set for its task
(\Cref{sec:appdx:baselines}). \emph{Self-Repair Gripper} is the only task with
DAgger corrections~\citep{ross2011dagger}: operator take-overs during ManiFlow
rollouts, each clipped into its own training episode
(\Cref{sec:appdx:baseline_deploy}). Its policies train on the last two rows
together. The Kitchen Organization set combines a rack-combo collection with a
multi-skill collection that omits the plate stage.}
\label{tab:appdx:real_data}
\vspace{4pt}
\centering
\small
\setlength{\tabcolsep}{8pt}
\renewcommand{\arraystretch}{1.15}
\begin{tabular}{@{}lrr@{}}
\toprule
\textbf{Task} & \textbf{Episodes} & \textbf{Hours} \\
\midrule
Put Plate on Rack                    & $301$   & $2.7$  \\
Sort Utensils                        & $152$   & $1.2$  \\
Kitchen Organization                 & $448$   & $3.8$  \\
Soft-Bag Zipping                     & $534$   & $9.7$  \\
Self-Repair Gripper                  & $802$   & $11.8$ \\
\midrule
Self-Repair Gripper (DAgger)         & $570$   & $5.6$  \\
\bottomrule
\end{tabular}
\end{table}

\textbf{Protocol.} We score every rollout under the partial-credit rubric of
\Cref{app:data-real-world-eval-criteria} and report the score normalized by the
maximum attainable score. Each task gets $20$ rollouts per method, except
\emph{Sort Utensils} ($10$). \emph{Put Plate on Rack} splits its $20$ into
one-plate and two-plate rubric variants of $10$ each. Object placements are re-randomized between
rollouts and methods are interleaved within each round, so every policy sees the
same lighting and environmental drift. Both \ours{} inference modes are runtime
flags on one fine-tuned checkpoint (\Cref{sec:appdx:yam}), not separate models.

\begin{figure}
    \centering
    \includegraphics[width=\linewidth]{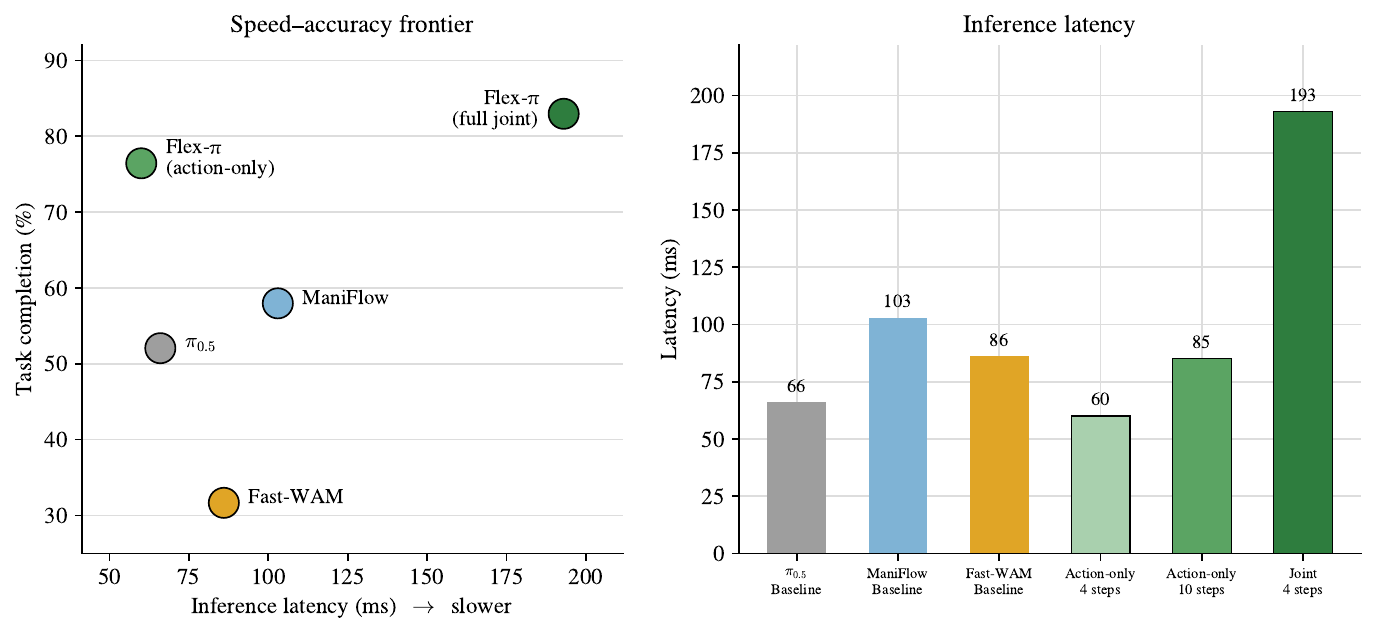}
    \caption{\textbf{Action-only is the fastest policy and also the most
    accurate; joint generation trades latency for further accuracy.}
    \emph{Left:} in-distribution task completion against single-inference
    latency on an RTX~5090, best stack per path
    (\Cref{tab:appdx:latency_joint,tab:appdx:latency_action}). The vertical
    axis is the five-task average reported in \Cref{sec:exp:real}, with
    \emph{Put Plate on Rack} pooled over its one- and two-plate rubric
    variants. \emph{Right:} the same latency measurements per configuration.
    At four denoise steps action-only is the fastest configuration shown,
    below all three baselines, and still delivers an $18.5\%$ gain over the
    strongest of them. Full joint generation costs roughly $3\times$ that
    latency and provides a further $6.5\%$ gain. Baselines run on only some of the
    five tasks are averaged over those.}
    \label{fig:real_perf_latency}
\end{figure}

\begin{figure}[t]
    \centering
    \includegraphics[width=0.92\linewidth]{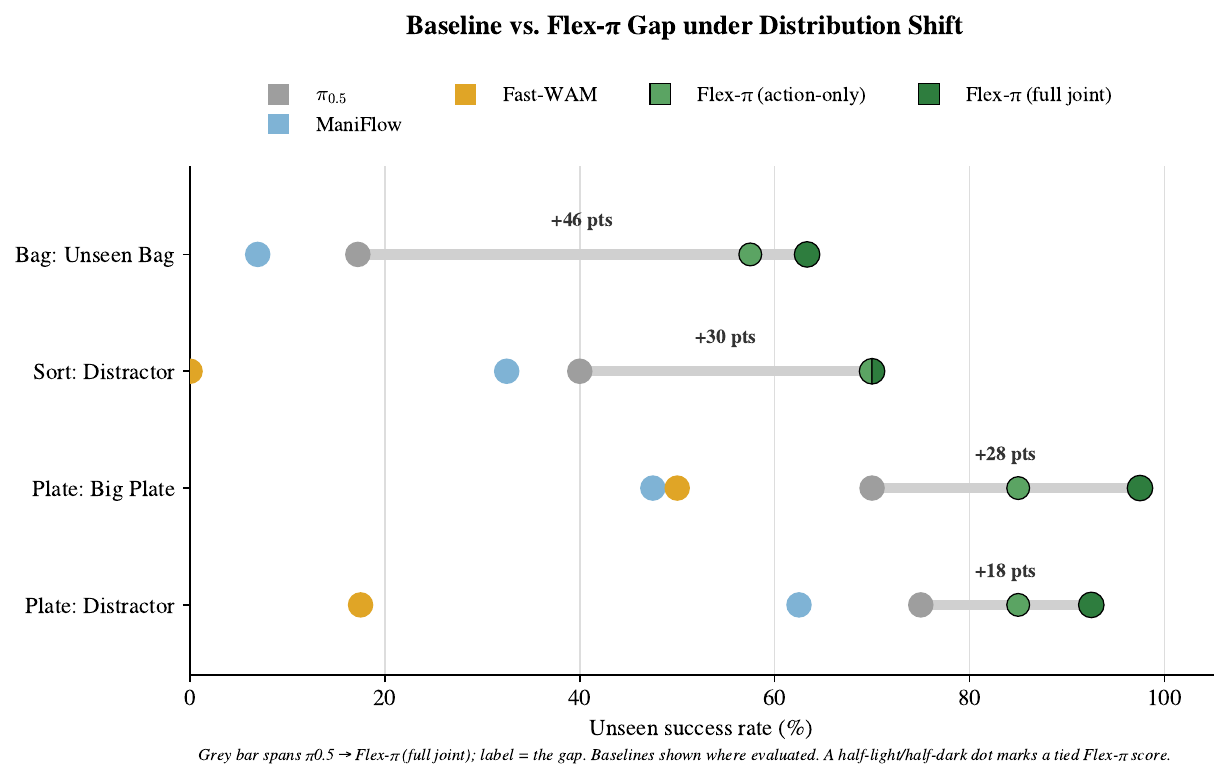}
    \caption{\textbf{\ours{} pulls further ahead of $\pi_{0.5}$ as the domain
    shift gets harder.} Each row is one unseen condition, ordered by the size of the
    gap; the grey bar spans $\pi_{0.5}$ to \ours{} (full joint). A half-light,
    half-dark dot marks a condition on which the two \ours{} settings score
    identically.}
    \label{fig:gap_unseen}
\end{figure}

\textbf{The margin over $\pi_{0.5}$ grows with the size of the shift.} \Cref{fig:gap_unseen}
orders the held-out conditions by the $\pi_{0.5}$-to-\ours{} success-rate
gain: $+18\%$ with added distractors on \emph{Put Plate on Rack}, $+28\%$ on
the unseen plate, $+30\%$ with added distractors on \emph{Sort Utensils}, and
$+46\%$ on the unseen soft bag. The ordering follows how much of the training-time appearance each
shift invalidates, matching the pattern LIBERO-Plus shows in simulation
(\Cref{tab:libero_plus}).

\begin{figure}[t]
    \centering
    \includegraphics[width=0.62\linewidth]{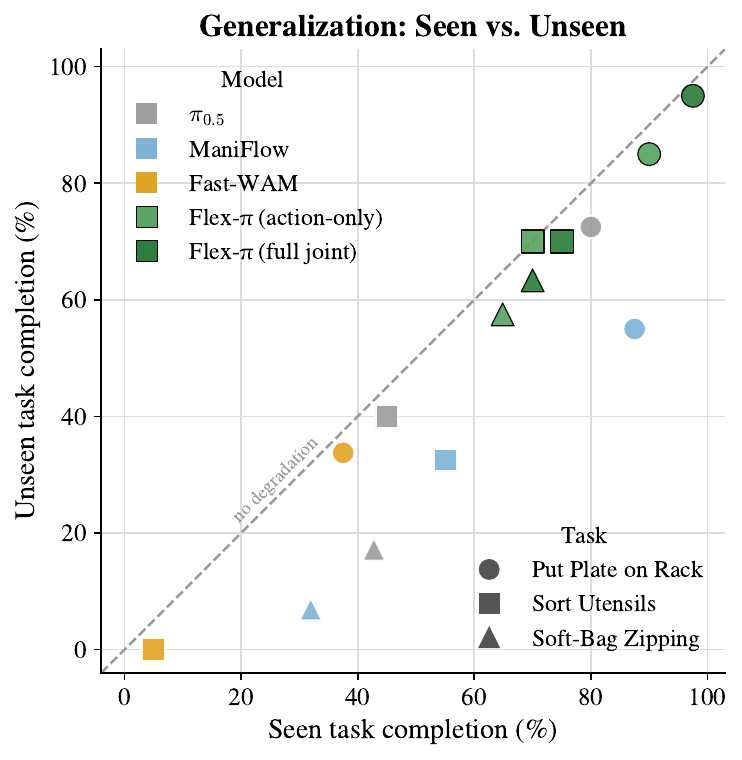}
    \caption{\textbf{How much each method loses under distribution shift.}
    Unseen task completion against the matching seen condition, one point per
    task; for \emph{Put Plate on Rack} the unseen coordinate averages the
    unseen-plate and distractor conditions. Distance below the dashed diagonal
    is the cost of the shift. Fast-WAM's full-completion rate is $0$\% in every
    condition of the two tasks it appears in, so both of its coordinates are
    partial credit for attempts that never finish.}
    \label{fig:seen_unseen_scatter}
\end{figure}

\textbf{\ours{} has the smallest drop from seen to unseen conditions.}
\Cref{fig:seen_unseen_scatter} plots each unseen condition against its matching
seen condition. Both \ours{} settings stay within $8\%$ of the diagonal on all three tasks,
and full joint stays within $7\%$ of the diagonal; their largest drop is on
the soft bag. ManiFlow starts from a comparable seen score on
\emph{Put Plate on Rack} and still exhibits a $22$--$33\%$ drop.
$\pi_{0.5}$ drops by less than $8\%$ on the two rigid-object tasks but by
$26\%$ on the soft bag, from a seen score $28\%$ below \ours{}.

\subsection{\ref{q4} Real-World Depth Input Ablation}
\label{sec:appdx:depth_ablation}

\noindent
\begin{minipage}[t]{0.615\linewidth}
\vspace{0pt}
\setlength{\parskip}{.5pc}
The pointmap carries metric geometry and is the only one of the three visual
streams \ours{} observes that needs a depth sensor at deployment. Withholding it
from the input costs nothing measurable. On \emph{Put Plate on Rack} in full
joint generation, task completion is $95.0\%$ with the depth input and
$91.7\%$ without (\Cref{fig:appdx:depth_ablation}).

Per-stream dropout with cross-modality forcing (\Cref{sec:method:dropout}) is
what makes this possible: the model is trained to generate each stream from the
others, so it can supply the geometry it is not given.
\Cref{fig:appdx:agibot_streams} shows that the generated scene structure
survives the same withholding. This makes the depth sensor optional at
deployment, but not the pointmap stream itself: removing it from \emph{training}
causes a $20.0\%$ performance drop in average RoboTwin success
(\Cref{fig:training_modalities}).
\end{minipage}
\hfill
\begin{minipage}[t]{0.35\linewidth}
\vspace{0pt}
\centering
\includegraphics[width=\linewidth]{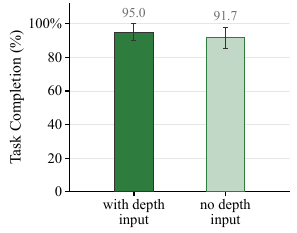}
\captionof{figure}{\textbf{Depth input is optional at deployment.} Task
completion on \emph{Put Plate on Rack}, with and without the depth input.}
\label{fig:appdx:depth_ablation}
\end{minipage}

\subsection{\ref{q1} Long-Horizon Dexterity: Self-Repair Gripper}
\label{sec:selfrepair}

\begin{figure}[t]
    \centering
    \begin{subfigure}[t]{0.48\linewidth}
        \centering
        \includegraphics[width=\linewidth]{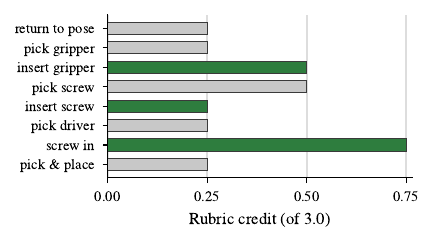}
        \caption{Rubric, in execution order.}
        \label{fig:selfrepair_stages}
    \end{subfigure}
    \hfill
    \begin{subfigure}[t]{0.48\linewidth}
        \centering
        \includegraphics[width=\linewidth]{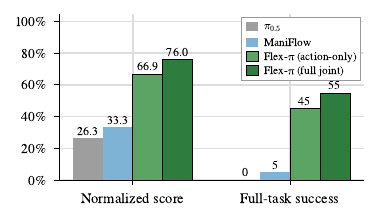}
        \caption{Performance.}
        \label{fig:selfrepair_results}
    \end{subfigure}
    \caption{\textbf{Self-Repair Gripper: an eight-stage task, and how far each method gets.}
    \textbf{Left:} the eight stages, which must be completed in order; green marks the three insertion and fastening stages, which together carry half of the $3.0$ available points.
    \textbf{Right:} partial-credit score normalized by the maximum attainable score, and the fraction of rollouts completing every stage.}
    \label{fig:selfrepair}
\end{figure}

In \emph{Self-Repair Gripper}, the robot repairs its own gripper, fastens it with a screw, and finally clears the workspace by placing a vegetable into a bucket. The task consists of eight sequential stages shared across both arms (\Cref{fig:selfrepair_stages}). The replacement gripper, screw holder, and screwdriver begin at randomized locations on the right side of the table, while the vegetable and bucket remain on the left. Three insertion and fastening stages account for half of the rubric's $3.0$ points, so high scores require accurate assembly rather than object transport alone.

\textbf{The three assembly stages have millimeter-level clearances.}
Seating the replacement gripper leaves $\pm0.5$\,mm of lateral clearance. Starting the screw is more forgiving, at $\pm1.75$\,mm, with the screw head preventing over-insertion. Driving the screw is the tightest stage, leaving only $\pm0.25$\,mm between the driver bit and the socket. Grasping the screwdriver automatically depresses its trigger, causing it to rotate for a fixed one-second interval, so failures arise from perception and placement rather than force control.

\textbf{Data collection and evaluation.}
We collect teleoperated demonstrations of the complete sequence at $30$\,Hz (\Cref{tab:appdx:real_data}). We then roll out a first-stage ManiFlow~\citep{yan2025maniflow} policy for two rounds and record human corrections wherever it fails; this is the only task with DAgger corrections. Using ManiFlow rather than \ours{} to collect corrections avoids biasing the dataset toward our own failure modes; ManiFlow is therefore also the only baseline evaluated on corrections generated from its own rollouts. All methods train on the identical dataset for a comparable number of steps. We evaluate \ours{} in both inference modes, ManiFlow, and $\pi_{0.5}$~\citep{intelligence2025pi05} over $20$ rollouts, reporting both normalized partial-credit score and complete-task success.

\textbf{Why we also report complete-task success.}
A policy that succeeds at each stage with $90\%$ probability still completes the
entire sequence only $43\%$ of the time. We therefore report complete-task
success alongside normalized partial credit.

\textbf{\ours{} leads every baseline, and full joint generation leads action-only.}
With full joint generation, \ours{} more than doubles the normalized score of the stronger baseline and completes the entire sequence in over half of its rollouts, where ManiFlow finishes one rollout in twenty and $\pi_{0.5}$ none (\Cref{fig:selfrepair_results}). Against action-only inference, full joint generation improves the normalized
partial-credit score by $9.1\%$ and complete-task success by $10\%$. Partial credit tolerates isolated failures whereas complete-task success does not, explaining why ManiFlow retains roughly a third of the rubric yet almost never finishes.

\subsection{\ref{q1} Deformable Manipulation: Soft-Bag Zipping}
\label{sec:softbag}

Unlike the previous tasks, \emph{Soft-Bag Zipping} requires manipulating an object with no stable rest shape. The robot unzips a fabric pouch, places one or two pens inside, and zips it shut. The pouch deforms whenever it is placed, opened, or loaded, so the geometry and location of the zipper pull vary across rollouts.

\textbf{The zipper pull requires precise localization and grasping.}
The pull is small, hangs slack, and is similar in color to the bag. The gripper must isolate the pull without catching the surrounding fabric, which can jam the slider. The rubric therefore scores grasping the pull separately from traversing the zipper (\Cref{app:data-real-world-eval-criteria}). Closing is the harder traverse because the loaded pouch deforms around its contents and the fabric must remain clear of the slider.

\textbf{Data collection and evaluation.}
We collect teleoperated demonstrations of the full sequence at $30$\,Hz (\Cref{tab:appdx:real_data}). Unlike Self-Repair Gripper, this task uses no DAgger corrections, so all methods train on the same demonstration set. The pouch starts beneath the head camera with a small randomized lateral displacement, while the pens are placed freely and picked with the nearer arm. We evaluate both inference modes of \ours{}, ManiFlow, and $\pi_{0.5}$~\citep{intelligence2025pi05} over $20$ rollouts each, reporting normalized partial credit and the fraction of rollouts that finish with the pouch zipped shut.

\textbf{\ours{} progresses through the deformable task where both baselines stall.}
Full joint generation scores over $1.5\times$ the stronger baseline and finishes the task twice as often as $\pi_{0.5}$ and eight times as often as ManiFlow (\Cref{fig:real_main}); action-only inference sits between full joint and the baselines on both measures. Because the rubric is sequential, the average raw score also indicates where rollouts tend to fail: both baselines stop before finishing the traverse that draws the slider open, whereas \ours{} in either mode gets past loading the pouch and stops short of the closing grasp.

This is also the only task on which $\pi_{0.5}$ outperforms ManiFlow in the seen
condition, after trailing it on the other four; under distribution shift, $\pi_{0.5}$ leads ManiFlow more broadly (\Cref{fig:seen_unseen_scatter}). One possible explanation is that dense 3D input provides less advantage for a deformable object whose geometry changes substantially across rollouts. Soft-Bag Zipping also shows a large gap between partial progress and completion: ManiFlow earns roughly one-third of the rubric on average but completes only one of twenty rollouts, echoing the gap observed over the longer sequence in \Cref{sec:selfrepair}.

\subsection{RoboTwin Ablation Setup}
\label{sec:appdx:robotwin_ablation}

The three RoboTwin ablations of \Cref{fig:robotwin_ablations_latency,fig:forcing_ablation} share one recipe: five tasks (\texttt{lift\_pot}, \texttt{place\_shoe}, \texttt{pick\_diverse\_bottles}, \texttt{place\_object\_basket}, and \texttt{stack\_bowls\_two}), $50$ demonstrations per task, $5$ epochs from scratch, evaluated under domain randomization. \Cref{fig:training_modalities,fig:forcing_ablation} train one model per variant on that recipe; \Cref{fig:latency_vs_success_robotwin} trains nothing further, varying only what a single checkpoint from that recipe generates at inference. Every variant sees the identical task set, demonstration budget, and schedule; only the ablated factor changes.

\subsection{Data Scaling Experiment in RoboTwin}
\label{sec:appdx:robotwin_scaling}

\textbf{\ours{} is the most data-efficient method at every demonstration
budget.} We re-train each method on fractional subsets of the RoboTwin
demonstrations and report success rate against dataset size
(\Cref{fig:robotwin scaling}). This experiment uses the $50$ random-scene
RoboTwin tasks, not the five-task recipe of the ablations above. At $50$ demos
per task \ours{} reaches $78.8\%$, against $31.4\%$ for
$\pi_{0.5}$~\citep{intelligence2025pi05}, $41.9\%$ for
Fast-WAM~\citep{yuan2026fast} and $17.2\%$ for
LingBot-VA~\citep{lingbot-va2026}. At $100$ demos it improves to $87.0\%$
against $44.7\%$, $68.1\%$ and $32.2\%$, and reaches $94.8\%$ at $500$ demos.

\subsection{Full LIBERO and LIBERO-Plus Results}
\label{sec:appdx:libero}
\Cref{tab:libero_flex} gives the full per-suite breakdown behind
\Cref{tab:libero_main}, and \Cref{tab:libero_plus} the per-perturbation
breakdown behind the LIBERO-Plus totals quoted in \Cref{sec:experiments}.

\begin{table}[t]
\caption{\textbf{LIBERO standard benchmark.} Per-suite and average success rate (\%); \textbf{bold} = best within each group (VLAs / WAMs). \ours{} is finetuned with flexible generation (one model, run action-only or jointly); \ours{}$^{*}$ is finetuned for a single fixed mode. \emph{no depth}: trained without the depth (pointmap) stream.}
\label{tab:libero_flex}
\vspace{4pt}
\centering
\small
\setlength{\tabcolsep}{4.5pt}
\renewcommand{\arraystretch}{1.15}
\begin{tabular}{@{}l|ccccc@{}}
\toprule
\multirow{2}{*}{\textbf{Methods}} & \multicolumn{5}{c}{\textbf{LIBERO}} \\
\cmidrule(lr){2-6}
 & \textbf{Spatial} & \textbf{Object} & \textbf{Goal} & \textbf{Long} & \textbf{Avg} \\
\midrule
\multicolumn{6}{@{}l}{\textit{VLAs}} \\
\midrule
Diffusion Policy~\citep{chi2023diffusion}                   & 78.3 & 92.5 & 68.3 & 50.5 & 72.4 \\
Octo~\citep{octo2024}                                       & 78.9 & 85.7 & 84.6 & 51.1 & 75.1 \\
OpenVLA~\citep{kim2024openvla}                              & 84.7 & 88.4 & 79.2 & 53.7 & 76.5 \\
SpatialVLA~\citep{qu2025spatialvla}                         & 88.2 & 89.9 & 78.6 & 55.5 & 78.1 \\
$\pi_0$+FAST~\citep{pertsch2025fast}                        & 96.4 & 96.8 & 88.6 & 60.2 & 85.5 \\
GR00T-N1~\citep{bjorck2025gr00t}                            & 94.4 & 97.6 & 93.0 & 90.6 & 93.9 \\
$\pi_0$~\citep{black2024pi0visionlanguageactionflowmodel}   & 96.8 & 98.8 & 95.8 & 85.2 & 94.1 \\
UniVLA~\citep{bu2025univla}                                 & 96.5 & 96.8 & 95.6 & 92.0 & 95.2 \\
$\pi_{0.5}$~\citep{intelligence2025pi05}                    & 98.8 & 98.2 & 98.0 & 92.4 & 96.9 \\
OpenVLA-OFT~\citep{kim2025oft}                              & 97.6 & 98.4 & 97.9 & 94.5 & 97.1 \\
RIPT-VLA~\citep{tan2025ript}                                & 98.6 & 98.6 & \textbf{99.0} & 93.8 & 97.5 \\
X-VLA~\citep{zheng2025xvla}                                 & 98.2 & 98.6 & 97.8 & 97.6 & 98.1 \\
MomloAct2~\citep{fang2026molmoact2}                         & 97.8 & \textbf{100.0} & 97.8 & 93.2 & 97.2 \\
MomloAct2-Think~\citep{fang2026molmoact2}                   & 98.8 & 99.8 & 98.5 & 95.4 & 98.1 \\
Qwen-RobotManip~\citep{yuan2026qwen}                        & --   & --   & --   & --   & 99.1 \\
Qwen-RobotManip-Context                                     & --   & --   & --   & --   & \textbf{99.2} \\
\ourcolor{\ours{} (action-only, no depth)}                  & 97.8 & 99.8 & 98.0 & 96.2 & 98.0 \\
\ourcolor{\ours{} (action-only)}                            & \textbf{99.4} & 99.6 & 98.0 & 96.6 & 98.4 \\
\ourcolor{\ours{}$^{*}$ (action-only, no depth)}            & 97.0 & \textbf{100.0} & 98.8 & \textbf{98.6} & 98.6 \\
\ourcolor{\ours{}$^{*}$ (action-only)}                      & 98.8 & \textbf{100.0} & 98.4 & 97.4 & 98.7 \\
\midrule
\multicolumn{6}{@{}l}{\textit{WAMs}} \\
\midrule
F1~\citep{lv2025f1}                                         & 98.2 & 97.8 & 95.4 & 91.3 & 95.7 \\
GE-Act~\citep{liao2025genie}                                & 98.2 & 97.6 & 95.8 & 94.4 & 96.5 \\
Fast-WAM~\citep{yuan2026fast}                               & 98.2 & \textbf{100.0} & 97.0 & 95.2 & 97.6 \\
Motus~\citep{bi2025motus}                                   & 96.8 & 99.8 & 96.6 & 97.6 & 97.7 \\
LingBot-VA~\citep{lingbot-va2026}                           & 98.5 & 99.6 & 97.2 & 98.5 & 98.5 \\
Cosmos-Policy~\citep{kim2026cosmospolicy}                   & 98.1 & \textbf{100.0} & 98.2 & 97.6 & 98.5 \\
\ourcolor{\ours{} (full joint, no depth)}                   & 99.0 & 99.8 & 98.6 & 96.6 & 98.5 \\
\ourcolor{\ours{} (full joint)}                             & \textbf{99.6} & 99.8 & 98.6 & 96.0 & 98.5 \\
\ourcolor{\ours{}$^{*}$ (full joint, no depth)}             & 98.8 & 99.6 & 98.6 & \textbf{98.8} & 99.0 \\
\ourcolor{\ours{}$^{*}$ (full joint)}                       & 99.2 & 99.8 & \textbf{99.0} & 98.6 & \textbf{99.2} \\
\bottomrule
\end{tabular}
\end{table}

\begin{table*}[t]
\caption{
\textbf{Robustness on LIBERO-Plus.}
Success rate (\%) under the seven perturbation types; \emph{Total} is the
official task-count-weighted mean over all $10{,}030$ perturbed tasks rather
than an unweighted mean of the seven categories.
\textbf{Bold} = best per column within each group (VLAs / WAMs).
The \ours{}, $\pi_{0.5}$, and Fast-WAM rows are our own evaluations under the
\emph{corrected} instruction protocol: the released LIBERO-Plus code
incorrectly builds the policy prompt from variant filenames for the six
non-language perturbation categories, appending perturbation metadata
(camera-viewpoint and initial-state identifiers) to instructions the benchmark
specifies as unchanged. Fixing the extraction restores each task's own language
annotation and leaves the rest of the protocol as released
(\Cref{sec:appdx:libero_setup}).
$^{\dagger}$$\pi_{0.5}$ and Fast-WAM publish no LIBERO-Plus results, so these
are our evaluations of their released checkpoints, with no weights updated. All
other baselines are published figures from \citet{fei25libero-plus}, which we
did not re-run.
The two \ours{} rows are one flexible checkpoint run under the two output masks,
not two separately fine-tuned models.
}
\label{tab:libero_plus}
\vspace{4pt}
\centering
\small
\setlength{\tabcolsep}{4.5pt}
\renewcommand{\arraystretch}{1.15}

\begin{tabular}{@{}l|ccccccc|c@{}}
\toprule
\textbf{Method}
& \textbf{Cam.}
& \textbf{Robot}
& \textbf{Lang.}
& \textbf{Light}
& \textbf{BG}
& \textbf{Noise}
& \textbf{Layout}
& \textbf{Total} \\
\midrule

\multicolumn{9}{@{}l}{\textit{VLAs}} \\
\midrule

OpenVLA~\citep{kim2024openvla}
& 0.8 & 3.5 & 23.0 & 8.1 & 34.8 & 15.2 & 28.5 & 15.6 \\

NORA~\citep{hung2025nora}
& 2.2 & 37.0 & 65.1 & 45.7 & 58.6 & 12.8 & 62.1 & 39.0 \\

UniVLA~\citep{bu2025univla}
& 1.8 & 46.2 & 69.6 & 69.0 & 81.0 & 21.2 & 31.9 & 42.9 \\

$\pi_0$~\citep{black2024pi0visionlanguageactionflowmodel}
& 13.8 & 6.0 & 58.8 & 85.0 & 81.4 & 79.0 & 68.9 & 53.6 \\

$\pi_0$-FAST~\citep{pertsch2025fast}
& 65.1 & 21.6 & 61.0 & 73.2 & 73.2 & 74.4 & 68.8 & 61.6 \\

RIPT-VLA~\citep{tan2025ript}
& 55.2 & 31.2 & 77.6 & 88.4 & 91.6 & 73.5 & 74.2 & 68.4 \\

OpenVLA-OFT~\citep{kim2025oft}
& 56.4 & 31.9 & 79.5 & 88.7 & 93.3 & 75.8 & 74.2 & 69.6 \\

Qwen-RobotManip~\citep{yuan2026qwen}
& 87.2 & 75.5 & 85.6 & 96.6 & 97.7 & 97.7 & 87.3 & 89.0 \\

Qwen-RobotManip-Context
& 89.9
& \textbf{83.9}
& \textbf{86.5}
& \textbf{98.6}
& \textbf{99.9}
& 97.9
& \textbf{87.5}
& \textbf{91.4} \\

$\pi_{0.5}$~\citep{intelligence2025pi05}$^{\dagger}$
& 76.0 & 75.9 & 85.8 & 96.9 & 95.9 & 88.9 & 86.6 & 85.7 \\

\ourcolor{\ours{} (action-only)}
& \textbf{91.6} & 70.1 & 83.3 & 98.0 & 97.4 & \textbf{98.1} & 86.4 & 88.6 \\

\midrule
\multicolumn{9}{@{}l}{\textit{WAMs}} \\
\midrule

Fast-WAM~\citep{yuan2026fast}$^{\dagger}$
& 48.4 & \textbf{75.7} & 66.8 & \textbf{96.6} & 68.5 & 70.1 & 76.5 & 70.8 \\

\ourcolor{\ours{} (full joint)}
& \textbf{89.1} & 72.7 & \textbf{84.5} & 94.6 & \textbf{96.3} & \textbf{98.9} & \textbf{87.3} & \textbf{88.6} \\

\bottomrule
\end{tabular}
\end{table*}
\clearpage
\section{Per-Task RoboTwin Results}
\label{sec:appdx:robotwin_pertask}
\Cref{tab:appdx:robotwin_pertask} gives the per-task breakdown behind the RoboTwin averages
of \Cref{tab:robotwin_main}, for the three baselines that report per-task figures and for
both \ours{} deployment modes.

\begin{table}[H]
\centering
\scriptsize
\setlength{\tabcolsep}{4pt}
\caption{\textbf{Per-task RoboTwin success rate (\%)} on all $50$ tasks under clean and
domain-randomized (\emph{Rand.}) evaluation, in the full-data setting of
\Cref{tab:robotwin_main}: $50$ clean $+$ $500$ randomized demonstrations per task
($2{,}500 + 25{,}000$ in total). Baseline columns are the published per-task figures of
\citet{yuan2026fast}; the \ours{} columns are our own evaluations of a single checkpoint in
its two deployment modes.}
\label{tab:appdx:robotwin_pertask}
\begin{tabular}{@{}l cc cc cc cc cc@{}}
\toprule
& \multicolumn{2}{c}{$\pi_{0.5}$}
& \multicolumn{2}{c}{Fast-WAM}
& \multicolumn{2}{c}{LingBot-VA}
& \multicolumn{2}{c}{\ourcolor{\ours{} (action-only)}}
& \multicolumn{2}{c@{}}{\ourcolor{\ours{} (full joint)}} \\
\cmidrule(lr){2-3}\cmidrule(lr){4-5}\cmidrule(lr){6-7}\cmidrule(lr){8-9}\cmidrule(l){10-11}
Task & Clean & Rand. & Clean & Rand. & Clean & Rand. & Clean & Rand. & Clean & Rand. \\
\midrule
adjust bottle & 100 & 99 & 100 & 100 & 90 & 94 & 100 & 100 & 98 & 97 \\
beat block hammer & 96 & 93 & 99 & 97 & 96 & 98 & 99 & 99 & 99 & 93 \\
blocks ranking rgb & 92 & 85 & 100 & 100 & 99 & 98 & 98 & 92 & 100 & 93 \\
blocks ranking size & 49 & 26 & 94 & 98 & 94 & 96 & 84 & 85 & 82 & 84 \\
click alarmclock & 98 & 89 & 100 & 100 & 99 & 100 & 100 & 98 & 91 & 91 \\
click bell & 99 & 66 & 100 & 100 & 100 & 100 & 98 & 95 & 94 & 97 \\
dump bin bigbin & 92 & 97 & 97 & 96 & 89 & 96 & 93 & 91 & 87 & 86 \\
grab roller & 100 & 100 & 100 & 100 & 100 & 100 & 100 & 100 & 100 & 100 \\
handover block & 66 & 57 & 95 & 81 & 99 & 78 & 100 & 96 & 99 & 95 \\
handover mic & 98 & 97 & 99 & 100 & 94 & 96 & 98 & 99 & 100 & 99 \\
hanging mug & 18 & 17 & 58 & 62 & 40 & 28 & 79 & 84 & 84 & 86 \\
lift pot & 96 & 85 & 100 & 100 & 100 & 99 & 100 & 100 & 99 & 100 \\
move can pot & 51 & 55 & 90 & 88 & 94 & 97 & 99 & 100 & 98 & 100 \\
move pillbottle pad & 84 & 61 & 100 & 99 & 99 & 99 & 100 & 100 & 100 & 100 \\
move playingcard away & 96 & 84 & 100 & 100 & 100 & 99 & 99 & 100 & 100 & 100 \\
move stapler pad & 56 & 42 & 77 & 64 & 91 & 79 & 93 & 92 & 96 & 94 \\
open laptop & 90 & 96 & 98 & 100 & 92 & 94 & 98 & 100 & 98 & 100 \\
open microwave & 34 & 77 & 62 & 45 & 82 & 86 & 60 & 64 & 70 & 77 \\
pick diverse bottles & 81 & 71 & 80 & 85 & 89 & 82 & 95 & 92 & 92 & 93 \\
pick dual bottles & 93 & 63 & 100 & 96 & 100 & 99 & 100 & 100 & 100 & 100 \\
place a2b left & 87 & 82 & 95 & 93 & 97 & 93 & 97 & 100 & 95 & 97 \\
place a2b right & 87 & 84 & 93 & 99 & 97 & 95 & 100 & 96 & 97 & 97 \\
place bread basket & 77 & 64 & 91 & 93 & 97 & 95 & 96 & 94 & 96 & 94 \\
place bread skillet & 85 & 66 & 90 & 93 & 95 & 90 & 90 & 90 & 90 & 95 \\
place burger fries & 94 & 87 & 96 & 99 & 97 & 95 & 98 & 97 & 98 & 99 \\
place can basket & 62 & 62 & 71 & 69 & 81 & 84 & 80 & 85 & 80 & 84 \\
place cans plasticbox & 94 & 84 & 99 & 96 & 100 & 99 & 100 & 100 & 100 & 100 \\
place container plate & 99 & 95 & 96 & 100 & 99 & 97 & 100 & 99 & 97 & 100 \\
place dual shoes & 75 & 75 & 94 & 88 & 94 & 89 & 93 & 93 & 98 & 93 \\
place empty cup & 100 & 99 & 100 & 100 & 100 & 100 & 100 & 100 & 100 & 100 \\
place fan & 87 & 85 & 96 & 96 & 99 & 93 & 96 & 96 & 97 & 97 \\
place mouse pad & 60 & 39 & 83 & 89 & 93 & 96 & 98 & 97 & 100 & 98 \\
place object basket & 80 & 76 & 89 & 88 & 91 & 88 & 90 & 92 & 84 & 92 \\
place object scale & 86 & 80 & 90 & 97 & 96 & 95 & 97 & 98 & 96 & 98 \\
place object stand & 91 & 85 & 90 & 94 & 99 & 96 & 96 & 98 & 97 & 99 \\
place phone stand & 81 & 81 & 97 & 99 & 97 & 97 & 97 & 98 & 98 & 100 \\
place shoe & 92 & 93 & 96 & 99 & 98 & 98 & 97 & 100 & 98 & 98 \\
press stapler & 87 & 83 & 90 & 97 & 85 & 82 & 96 & 99 & 97 & 98 \\
put bottles dustbin & 84 & 79 & 95 & 90 & 87 & 91 & 96 & 95 & 95 & 96 \\
put object cabinet & 80 & 79 & 94 & 89 & 85 & 87 & 81 & 84 & 90 & 95 \\
rotate qrcode & 89 & 87 & 93 & 89 & 96 & 91 & 91 & 89 & 84 & 88 \\
scan object & 72 & 65 & 89 & 92 & 96 & 91 & 91 & 89 & 92 & 87 \\
shake bottle & 99 & 97 & 100 & 100 & 100 & 97 & 100 & 100 & 100 & 99 \\
shake bottle horizontally & 99 & 99 & 100 & 100 & 100 & 99 & 100 & 100 & 100 & 99 \\
stack blocks three & 91 & 76 & 95 & 97 & 99 & 98 & 99 & 98 & 99 & 96 \\
stack blocks two & 97 & 100 & 100 & 100 & 100 & 98 & 100 & 100 & 100 & 100 \\
stack bowls three & 77 & 71 & 80 & 81 & 86 & 83 & 85 & 85 & 77 & 84 \\
stack bowls two & 95 & 96 & 92 & 98 & 94 & 98 & 93 & 96 & 95 & 98 \\
stamp seal & 79 & 55 & 90 & 94 & 96 & 97 & 98 & 100 & 99 & 99 \\
turn switch & 62 & 54 & 61 & 59 & 44 & 45 & 77 & 77 & 81 & 77 \\
\midrule
\textit{Average} & 82.7 & 76.8 & 91.9 & 91.8 & 92.9 & 91.5 & 94.5 & 94.6 & 94.3 & 94.8 \\
\bottomrule
\end{tabular}
\end{table}

\section{Inference Optimization}
\label{sec:appdx:inference}
 
The latencies quoted throughout the paper, including \Cref{tab:robotwin_steps}, are
measured on the deployment stack described here. Every component is
\emph{training-free}: no distillation, no architectural change, no dropped stream, and
the same checkpoint throughout, so the success rates reported elsewhere in the paper
transfer unchanged, subject to the fidelity checks below.

\textbf{Protocol.}
All measurements use one NVIDIA RTX~5090 ($32$\,GB), PyTorch~2.7.1/CUDA~12.8 with TensorRT~10.16, and the same fine-tuned checkpoint. Inputs use the deployed three-camera $384{\times}320$ composite, including RGB, depth, intrinsics, and proprioception. The $128$-token language context is precomputed and cached per task. Each configuration runs in a separate process with $3$ warmup and $20$ timed calls, synchronizing after each call; we report the mean. All runs use Euler integration.

\textbf{Token accounting.} At full joint generation, the model processes $360$ video $+441$ DINO $+360$ pointmap $+32$ action $=1193$ tokens across $30$ blocks, including $387$ first-frame anchors and $806$ noisy visual/action tokens. For action-only generation, the same $387$ anchors are prefetched into a key/value cache, and each step denoises only the $32$ action tokens. Generated visual latents are not copied back to the host, since deployment only consumes the actions. This saves $\sim25$\,ms per call in the full-generation setting.

\textbf{Joint path.} \Cref{tab:appdx:latency_joint} is a ladder in which each row differs
from the one above by exactly one component. Fitting
$T(K) = \text{fixed} + K \cdot \text{per-step}$ over the four step counts separates
one-per-call from one-per-step cost, and each component falls cleanly into one term or
the other. Exporting the $30$-block denoise core to a TensorRT engine is purely
per-step, cutting it by $48\%$; the dense joint attention mask rides as a runtime input
rather than a baked constant, so one engine serves every input/output regime. The two
host-side components remove CPU work and nothing else, and so move only the fixed term.
Only whole-loop compilation touches both, since capturing the loop as one graph fuses
kernels and collapses per-call launch overhead at once. The best stack is therefore
step-dependent, and the achievable speedup rises from $2.0\times$ at $K{=}1$ to
$2.5\times$ at $K{=}10$: engine work attacks the term that scales with $K$, while the
${\sim}20$\,ms floor of encoders and host glue does not.

\begin{table}[t]
\caption{\textbf{Joint-path inference ladder} (ms/call, RTX~5090, full input and full
joint generation). Each row adds one component to the row above except where noted.
\emph{fixed} and \emph{per-step} are the least-squares decomposition of $T(K)$ over the
four step counts.}
\label{tab:appdx:latency_joint}
\vspace{4pt}
\centering
\small
\setlength{\tabcolsep}{5pt}
\renewcommand{\arraystretch}{1.15}
\begin{tabular}{@{}ll|cccc|cc@{}}
\toprule
& \textbf{Stack} & $K{=}1$ & $K{=}2$ & $K{=}4$ & $K{=}10$ & \textbf{fixed} & \textbf{per-step} \\
\midrule
L0 & eager \texttt{bf16}, SDPA (baseline)            & 147.2 & 246.0 & 447.5 & 1065.3 & 42.2 & 102.2 \\
L1 & $+$ SDPA backend auto-selection                 & 152.7 & 251.3 & 447.9 & 1053.3 & 50.6 & 100.2 \\
L2 & $+$ loop-scope compile $+$ compiled encoders    & 106.3 & 191.2 & 359.5 & 867.8  & 21.6 & 84.6  \\
L3 & TensorRT joint core \emph{(replaces L2)}        & 97.1  & 148.7 & 252.0 & 572.0  & 42.9 & 52.8  \\
L4 & $+$ host-glue memoization                       & 79.2  & 131.0 & 236.3 & 553.4  & 25.9 & 52.7  \\
L5 & $+$ encoder CUDA graphs                         & \textbf{73.1} & 125.5 & 230.0 & 545.5 & 20.4 & 52.5 \\
L6 & prefill/decode split \emph{(replaces L3)}, $+$L4, $+$L5 & 77.8 & \textbf{115.9} & \textbf{193.3} & \textbf{428.1} & ${\sim}20$ & 38.5 \\
\bottomrule
\end{tabular}
\end{table}

\textbf{Prefill/decode split.} The $387$ anchor tokens attend only to one another and are
modulated at flow time $\tau{=}0$, so their per-layer keys and values are step-invariant;
we verify this directly, as the anchor output of the joint block stack is bit-identical
across denoise steps. Prefilling them once and decoding only the $806$ noisy tokens
against the cached prefix therefore removes ${\sim}32\%$ of the tokens from every step
without changing steps, streams, or math, and cuts per-step cost a further $27\%$. As a
TensorRT engine the decode/full ratio reaches $0.719$ against a $0.676$ token ratio, the
gap being the key/value assembly the engine fuses and the eager path pays for. The split
is bought with a ${\sim}20$\,ms once-per-call prefill, so it pays only for $K \ge 2$; at
$K{=}1$ the single engine wins by ${\sim}5$\,ms. One caveat for anyone re-measuring it:
the prefill is cached on anchor content, so a benchmark replaying one fixed observation
pays it once and reports the split ${\sim}20$\,ms/call too fast. The L6 row of
\Cref{tab:appdx:latency_joint} carries the correction, measured by counting engine
launches under rotating versus static observations ($20$ versus $0$ prefills over $20$
calls); the single-engine stack has no such cache and is flat under the same test.

\textbf{Action-only path.} The fast path inverts the picture
(\Cref{tab:appdx:latency_action}): the dominant lever is loop-scope compilation, not
TensorRT. A $32$-token step against a cached prefix is launch-bound rather than
FLOP-bound, so capturing the entire loop, including the key/value prefill, as one graph
cuts per-step cost $3.5\times$, from $14.2$ to $4.1$\,ms, far more than better kernels
could. The two host-side components reproduce their joint-path effects almost exactly, as
expected for costs that are per-call and independent of what is being denoised.

\begin{table}[t]
\caption{\textbf{Action-only inference ladder} (ms/call, RTX~5090, full input, no visual
stream generated). Columns as in \Cref{tab:appdx:latency_joint}.}
\label{tab:appdx:latency_action}
\vspace{4pt}
\centering
\small
\setlength{\tabcolsep}{5pt}
\renewcommand{\arraystretch}{1.15}
\begin{tabular}{@{}ll|cccc|cc@{}}
\toprule
& \textbf{Stack} & $K{=}1$ & $K{=}2$ & $K{=}4$ & $K{=}10$ & \textbf{fixed} & \textbf{per-step} \\
\midrule
A0 & eager \texttt{bf16}, SDPA (baseline)         & 89.1 & 100.4 & 131.8 & 215.7 & 73.9 & 14.2 \\
A1 & $+$ SDPA backend auto-selection              & 89.3 & 102.9 & 128.5 & 220.3 & 73.1 & 14.6 \\
A2 & $+$ host-glue memoization $+$ encoder graphs & 66.7 & 80.1  & 106.7 & 188.7 & 52.9 & 13.6 \\
A3 & loop-scope compile \emph{(replaces eager)}   & 56.0 & 60.1  & 66.8  & 92.9  & 51.5 & 4.1  \\
A4 & $+$ compiled encoders                        & 49.0 & \textbf{52.6} & \textbf{60.3} & \textbf{85.2} & 44.6 & 4.0 \\
A5 & $+$ encoder CUDA graphs instead              & \textbf{48.6} & 52.5 & 62.0 & 87.4 & 44.2 & 4.3 \\
\bottomrule
\end{tabular}
\end{table}

\textbf{What did not work.} Several standard accelerations are no-ops on this stack.
FP8 inside the TensorRT engine is a $1.4\%$ wash with slightly worse numerics, since on
this architecture the FP16 tactics already saturate, and MXFP8/NVFP4 engines do not build
at all under TensorRT~$10.16$. A whole-loop CUDA graph over the engine path is impossible,
as the engine's async execution is not capturable, and would have bought only
${\sim}3\%$. Compiling the encoders is a no-op on the \emph{joint} engine path, as is
overlapping the three encoders on separate streams, for the same root cause: those
encoders are launch-bound, and four per-forward synchronizing constants broke both graph
capture and compilation until removed. Raising the TensorRT builder optimization level
from~$3$ to~$5$ changes per-step cost by $0.5\%$. A second-order multistep solver is
mathematically sound but collapses in rollout ($20.5\%$ vs $94.7\%$ success at six
steps): these checkpoints are Euler-tuned, and plain Euler with fewer steps is the better
trade.

\textbf{What imagination costs.} With the same checkpoint, the same observation, and the
best stack for each path, the only difference between the two regimes is whether the
three visual streams are denoised. At $K{=}1$ they nearly converge ($49.0$ vs
$73.1$\,ms, $1.5\times$); at $K{=}10$ they differ by $5\times$ ($85.2$ vs $428.1$\,ms).
The ratio grows because imagination is almost entirely per-step: $38.5$\,ms for the
$1193$-token joint sequence against $4.0$\,ms for $32$ action tokens, over a shared
${\sim}20$\,ms floor. Cutting steps is therefore the one lever that shortens the joint
path without giving up a stream (\Cref{tab:robotwin_steps}).

\textbf{Numerical fidelity.} Host-glue memoization and encoder CUDA graphs are bit-exact
by construction, since they remove host work and nothing else, and we verify this with
identical fixed-seed action fingerprints. The TensorRT engine does introduce rounding:
$1.4\%$ relative $L_2$ per denoise step, but only $0.55\%$ on the final action chunk, so
the denoise loop contracts per-step error rather than compounding it. We validated the
engine end-to-end separately, where it ties the compiled PyTorch path at $90.7\%$ success
over three RoboTwin tasks $\times$ $50$ episodes. The $K{=}1$ column of
\Cref{tab:appdx:latency_joint} is a latency floor rather than an operating point: success
collapses there (\Cref{tab:robotwin_steps}).

\textbf{Memory.} Torch's allocator does not see TensorRT engine weights or scratch, so
torch-side figures understate the engine stacks; sampling the whole process instead gives
$15.8$\,GB peak for the eager and compiled stacks, $26.4$\,GB for the single engine, and
$25.7$\,GB for the prefill/decode split. The two engine stacks cost essentially the same
resident memory despite the split's on-disk footprint being ${\sim}9$\,GB larger, because
the split shares one scratch buffer sized to the larger of its two engines and both
stacks free the now-dead PyTorch expert weights once the engines are installed. On a
$32$\,GB card this leaves $5$--$6$\,GB of headroom: enough for pure inference, but tight
when a simulator's renderer is co-resident.

\clearpage
\subsection{Denoising Steps}
\label{sec:appdx:steps}

\begin{wraptable}{r}{0.35\textwidth}
\vspace{-\intextsep}
\centering
\small
\renewcommand{\arraystretch}{1.15}
\begin{tabular}{@{}c|ccc@{}}
\toprule
$K$ & \textbf{Latency} & \textbf{Clean} & \textbf{Rand.} \\
\midrule
10 & 85 & 93.5 & 93.6 \\
4  & 60 & \textbf{94.5} & \textbf{94.6} \\
2  & 53 & 93.7 & 93.9 \\
1  & 49 & 51.0 & 52.9 \\
\bottomrule
\end{tabular}
\caption{\textbf{Denoising steps, action-only.} RoboTwin success (\%) and latency (ms, RTX~5090, measured on the stack of \Cref{sec:appdx:inference}) from one checkpoint. Success peaks at $K{=}4$.}
\label{tab:robotwin_steps}
\end{wraptable}

The number of Euler steps $K$ is a deploy-time knob orthogonal to the choice of streams.
Joint generation denoises every active stream at every step, so its cost is near-linear in $K$; the action-only path denoises only the $32$ action tokens against a cached prefix, so $K$ costs it far less.
Sweeping $K$ on the same checkpoint in action-only mode (\Cref{tab:robotwin_steps}), success peaks at $K{=}4$ and stays within $1.0$ point of that peak for every $K \ge 2$, then collapses below $60\%$ at a single step.
Predicting in latent space rather than pixels is what buys this: the action expert reads the future streams for geometry and semantics, which settle well before the latents are visually converged.
Full joint generation was swept over the same $K$ and peaks at $K{=}4$ as well, so we use $K{=}4$ in both modes for the main simulation and real-world experiments.

\section{Training Hyperparameters}
\label{sec:appdx:experiments}

\Cref{tab:appdx:hparams} lists the settings shared by every run reported in this paper.
Pre-training and each fine-tuning domain differ only in the dataset, the number of
epochs, and the batch size; the optimizer, schedule, precision, and all stream-specific
settings are held fixed, so a comparison across domains is a comparison of data rather
than of recipe. Architectural sizes (expert widths, depth, head counts, adapter
construction) are given in \Cref{sec:appdx:architecture} and are not repeated here.

\textbf{What is trained.} The Wan-2.2 VAE, the \texttt{umT5} text encoder, and the
DINOv3 encoder are frozen throughout pre-training and fine-tuning. We train the shared
visual trunk, the action expert, the per-stream adapters, and the per-stream output
heads. The action expert is initialized by resampling the Wan-2.2 blocks
(\Cref{sec:appdx:adapters}); everything else in the trunk is loaded from
\texttt{Wan-2.2-5B}, and only the action encoder and action head start from scratch.

\textbf{Stream dropout.} The two masks of \Cref{sec:method:dropout} are drawn per sample
from the Bernoulli probabilities in \Cref{tab:appdx:hparams}, independently of one
another, with rejection sampling on the input mask so that at least one visual stream is
always observed (\Cref{sec:appdx:regimes}). Cross-modality forcing is enabled for all
three visual streams, so a stream dropped from the input is still denoised at the output.

\begin{table}[t]
\caption{\textbf{Training hyperparameters.} Shared across AGIBOT World pre-training and
all fine-tuning runs unless noted. Per-domain values are given in the last group.}
\label{tab:appdx:hparams}
\vspace{4pt}
\centering
\small
\setlength{\tabcolsep}{6pt}
\renewcommand{\arraystretch}{1.1}
\begin{tabular}{@{}ll@{}}
\toprule
\textbf{Setting} & \textbf{Value} \\
\midrule
\multicolumn{2}{@{}l}{\textit{Optimization}} \\
\quad Optimizer & AdamW, $\beta = (0.9,\, 0.95)$ \\
\quad Learning rate & $1 \times 10^{-4}$ \\
\quad Schedule & cosine, linear warmup over the first $5\%$ of steps \\
\quad Weight decay & $1 \times 10^{-2}$ \\
\quad Gradient clipping & $1.0$ (global norm) \\
\quad Precision & \texttt{bfloat16} mixed precision \\
\quad Sharding & DeepSpeed ZeRO stage~1 via Accelerate \\
\midrule
\multicolumn{2}{@{}l}{\textit{Sample window} (\Cref{sec:appdx:actions})} \\
\quad Timesteps per sample & $33$ \\
\quad Action chunk & $H = 32$ \\
\quad Visual frames per sample & $9$ (stride $4$) \\
\midrule
\multicolumn{2}{@{}l}{\textit{Streams}} \\
\quad Loss weights $\lambda_a, \lambda_o, \lambda_d, \lambda_p$ & $1.0$ each \\
\quad Flow-matching shift (video / DINO / pointmap) & $6.0$ \\
\quad Flow-matching shift (action) & $1.0$ \\
\quad DINO encoder & DINOv3 ViT-B/16, frozen, $768$-d \\
\quad Pointmap encoding & shared VAE, clipped at $2$\,m \\
\quad Language context & $128$ \texttt{umT5} tokens $+\ 1$ proprioception token \\
\midrule
\multicolumn{2}{@{}l}{\textit{Per-sample stream dropout} (\Cref{sec:method:dropout})} \\
\quad $p(\text{stream observed})$ & $0.5$ per visual stream \\
\quad $p(\text{stream read by the action tokens})$ & $0.5$ per visual stream \\
\quad Constraint on the input mask & at least one visual stream observed \\
\quad Cross-modality forcing & enabled for all three visual streams \\
\bottomrule
\end{tabular}
\end{table}

\end{document}